\pdfoutput=1

\documentclass[11pt]{article}

\usepackage[preprint]{acl}
\usepackage{booktabs}%
\usepackage{pifont}
\usepackage{times}
\usepackage{array}
\usepackage{latexsym}
\usepackage{multirow}
\usepackage[table]{xcolor}

\usepackage[T1]{fontenc}

\usepackage[utf8]{inputenc}
\usepackage{float}
\usepackage{stfloats}  
\usepackage{enumitem}
\usepackage{url}
\usepackage{makecell}
\newcolumntype{C}[1]{>{\centering\arraybackslash}m{#1}}

\usepackage{microtype}

\usepackage{inconsolata}
\usepackage{amsmath}
\usepackage{graphicx}

\title{NovBench: Evaluating Large Language Models on Academic Paper Novelty Assessment}

\author{
 \textbf{Wenqing Wu\textsuperscript{1,2}},
 \textbf{Yi Zhao\textsuperscript{3}},
 \textbf{Yuzhuo Wang\textsuperscript{3}},
 \textbf{Siyou Li\textsuperscript{2}},
\\
 \textbf{Juexi Shao\textsuperscript{2}},
 \textbf{Yunfei Long\textsuperscript{2 \dag}},
 \textbf{Chengzhi Zhang\textsuperscript{1 \dag}}
\\
\\
 \textsuperscript{1}School of Economics and Management, Nanjing University of Science and Technology,\\
 \textsuperscript{2}School of Electronic Engineering and Computer Science, Queen Mary University of London,\\
 \textsuperscript{3}School of Management, Anhui University
\\
 \small{
   \textbf{\dag Correspondence:} \href{mailto:zhangcz@njust.edu.cn}{zhangcz@njust.edu.cn}, \href{mailto:yunfei.long@qmul.ac.uk}{yunfei.long@qmul.ac.uk} 
 }
}

\begin{document}
\maketitle
\begin{abstract}
Novelty is a core requirement in academic publishing and a central focus of peer review, yet the growing volume of submissions has placed increasing pressure on human reviewers. While large language models (LLMs), including those fine-tuned on peer review data, have shown promise in generating review comments, the absence of a dedicated benchmark has limited systematic evaluation of their ability to assess research novelty. To address this gap, we introduce NovBench, the first large-scale benchmark designed to evaluate LLMs' capability to generate novelty evaluations in support of human peer review. NovBench comprises 1,684 paper–review pairs from a leading NLP conference, including novelty descriptions extracted from paper introductions and corresponding expert-written novelty evaluations. We focus on both sources because the introduction provides a standardized and explicit articulation of novelty claims, while expert-written novelty evaluations constitute one of the current gold standards of human judgment. Furthermore, we propose a four-dimensional evaluation framework (including Relevance, Correctness, Coverage, and Clarity) to assess the quality of LLM-generated novelty evaluations. Extensive experiments on both general and specialized LLMs under different prompting strategies reveal that current models exhibit limited understanding of scientific novelty, and that fine-tuned models often suffer from instruction-following deficiencies. These findings underscore the need for targeted fine-tuning strategies that jointly improve novelty comprehension and instruction adherence.
\end{abstract}

\section{Introduction}
Novelty is a fundamental aspect of publication decisions in academic research, requiring that a paper's content or methodology makes a meaningful contribution to advancing existing knowledge, rather than simply replicating or validating established findings \citep{novelty_intro}. As the primary quality control mechanism in scientific research, assessing novelty is one of the core functions of the peer review system \citep{peer_intro, intro2}. However, the peer review process is currently facing pressure due to the explosive growth in academic submissions \citep{sub_increase} and the widening gap in the availability of qualified reviewer resources \citep{reviewer_resource}. This pressure directly manifests as a significant challenge for the robust evaluation of a paper's novelty \citep{Zhao}. Although methods for novelty evaluation have been proposed based on bibliometric \citep{uzzi,Matsumoto,Shibayama} and deep learning \citep{dl_nov1,dl_nov2,dl_nov3} approaches, these techniques predominantly focus on quantitative metrics (or numerical indicators). Compared to the textual evaluations provided by reviewers, these quantitative metrics inherently lack the necessary interpretability, making it difficult to effectively assist reviewers in their judgment during the peer review process or to provide authors with targeted advice for manuscript improvement.

Large Language Models (LLMs) have demonstrated exceptional capabilities across a wide range of scientific tasks \citep{2024autosurvey,bao} and show potential in assisting academic peer review \citep{zhou,llm4pr2}. Recent researches have attempted to enhance the capabilities of LLMs in automated peer review through various avenues, such as applying more effective prompting strategies \citep{liang1}, fine-tuned model \citep{finetune1,finetune2}, and multi-agent frameworks \citep{agent1,agent2}. 
Despite the promising progress demonstrated by recent advances, several critical issues remain insufficiently explored. Although these methods achieve competitive performance on paper-level scoring tasks, the evaluation of generated review text itself has received relatively little attention. Existing evaluation approaches are largely limited to metrics such as ROUGE \citep{rouge}, BLEU \citep{bleu}, and BERTScore \citep{BERTScore}, or the adoption of the LLM-as-judge \citep{llmasjudge}. However, these methods either rely on surface-level lexical similarity or depend on non-transparent LLM-based judgments, and therefore fail to reliably assess the semantic adequacy and aspect-specific correctness required for evaluating free-form novelty evaluations \citep{kuznetsov2024natural}. Furthermore, existing research primarily focuses on generating holistic peer review text, treating the review as a monolithic output. As a result, the ability of both general purpose and fine tuned models to perform aspect specific evaluation, particularly novelty evaluation, remains poorly characterized. Without isolating novelty as an independent evaluation target, it is difficult to determine whether a model genuinely evaluates research novelty or merely generates fluent, plausible sounding review like language. Therefore, the performance of existing general and fine-tuned models on the novelty evaluation task constitutes an important area for further investigation.

To address these issues, we present a novel evaluation framework, to be more specific: (1) We introducing a structured benchmark for novelty evaluation. This resource, which incorporates the' textual evaluations of the novelty dimension of the reviewers alongside the novelty descriptions of the paper introductions, constitutes a critical resource for future research. (2) We propose a four-dimensional, interpretable, and semantics-aware framework for evaluating LLM-generated novelty review text. This metric suite surpasses the limitations of traditional lexical overlap metrics (e.g., ROUGE/BLEU) and more effectively captures the quality of "novelty" evaluation in free-form text. (3) We conduct systematic evaluation and benchmarking of current general LLMs (e.g., $\text{GPT-5}$ \citep{gpt5}, $\text{Gemini-2.5-flash}$ \citep{gemini2.5}) and specialized LLMs on the novelty evaluation task. Based on this benchmark, we further analyze the gap between LLM-generated evaluations and human judgments, providing an in-depth understanding of these models' advantages and limitations in identifying and articulating novelty. (4) We conduct a comprehensive empirical analysis that identifies key factors for generating high-quality, highly interpretable novelty review text, and reveals important behavioral patterns of LLMs in novelty evaluation, thus directing the future development of more reliable AI-assisted peer review. \footnote{All resources publicly available at \url{https://github.com/njust-winchy/llm4novelty}.}
\section{Related Work}
Automated Scholarly Paper Review (ASPR) \citep{LIN2023101830} refers to the process in which computers or intelligent machines independently evaluate the content of a scholarly paper and generate a review report automatically. Early research concerning ASPR predominantly focused on paper rating recommendation \citep{rw1,rw2,rw3}. Furthermore, some studies \citep{rw4,rw5} have attempted to fine-tune pre-trained models to generate paper reviews.

With the LLMs demonstrate powerful text generation capabilities \citep{rw6,rw7}, the application of LLMs to help peer review has rapidly become a significant research focus. Numerous studies have evaluated or benchmarked the performance of LLMs in generating reviews of academic papers \citep{zhou,liang1,du}. The results of these investigations indicate that while LLMs are capable of providing meaningful feedback, they often lack critical analysis, and the comments generated frequently lack the insights and specificity found in human-written reviews.

Consequently, the research community has devoted significant efforts for enhancing LLM performance in review generation \citep{SEA,reviewer2,openreviewer,cycleresearcher,deepreviewer,treereview}. However, recent advancements have broadly evaluated and improved the overall performance of LLMs in review generation. Despite this progress, their efficacy in assessing specific, fine-grained aspects of a paper (especially novelty) remains an area requiring further investigation.

Novelty is recognized as a key aspect for measuring the quality and contribution of academic papers. Concurrently, as LLMs continue to demonstrate increasing capability, researchers have advanced their investigation into the role of LLMs in scientific novelty, including LLM-based assessment of paper novelty \citep{dl_nov1,dl_nov3,novelty1,novelty3,wu1,wu2,novelty2} and the generation of novel research ideas \citep{idea1,idea2,idea3,graphmind}.

Although these studies suggest that LLMs possess a certain capacity for novelty assessment, the evaluation of LLM performance is largely centered on quantitative output scores or relies on human evaluation. There remains a notable absence of dedicated assessment targeting the LLM generated textual novelty evaluations themselves, with current practices frequently defaulting to the "LLM as a judge" paradigm. This reliance exposes a gap in the development of robust evaluation methodologies for free form assessment text. While prior work has introduced resources for analyzing research limitations \citep{limit}, there is currently no publicly available resource dedicated to novelty assessment. To address this gap, we introduce a dedicated resource for research novelty and a novel methodology for assessing novelty evaluation. This paper deliberately focuses on novelty, as it is widely regarded as a central criterion for publication decisions and one of the most conceptually challenging aspects to evaluate \citep{beyond}.
\section{NovBench}
\begin{figure*}[t]
  \includegraphics[width=1.0\linewidth]{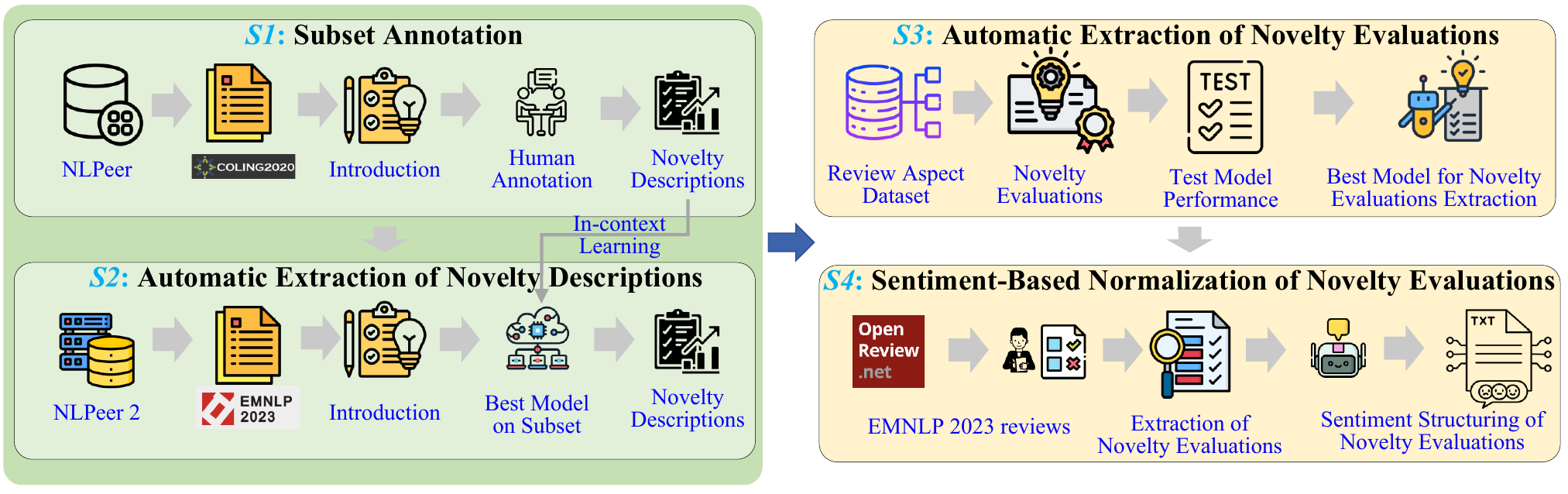}
  \caption {The pipeline for constructing NovBench, consisting of four stages.}
  \label{overview}
\end{figure*}
\subsection{Dataset Construction}
We initially collected the COLING 2020 and EMNLP 2023 datasets (include parsed paper in intertextual graphs (ITG) format \citep{itg} and review report) from NLPeer \citep{r1} and OpenReview\footnote{https://openreview.net/}, followed by the execution of three automated stages: (1) extraction of novelty descriptions from the paper introductions, (2) extraction of novelty-related evaluations from the reviewer comments, and (3) structuring novelty evaluations based on sentiment polarity. We extract novelty descriptions from paper introductions because they provide the most explicit and standardized articulation of authors’ claimed contributions \citep{evi2,evi1}. Figure \ref{overview} shows the construction workflow. 
\subsubsection{Automatic Extraction of Novelty Descriptions from Introduction}
We first extracted all content under the heading "Introduction" from the parsed paper. Subsequently, we adopted a strategy of annotating a small subset (COLING 2020, 87 papers) of the data to evaluate the performance of LLMs, which was then applied to the large dataset (EMNLP 2023, 1,684 papers). Specifically, we manually annotated the novelty descriptions extracted from the introductions of the COLING 2020 papers. We then designed various prompts to evaluate the performance of mainstream LLMs on this specific task, the detailed results are presented in the Appendix \ref{novelty_desc}. Finally, we selected GPT-5 with in-context-learning prompt as the method for the automatic extraction of novelty descriptions for the introduction of EMNLP 2023.

\subsubsection{Automatic Extraction of Novelty Evaluations from Peer Review Texts}
To extract novelty related evaluations from reviewer comments, it is necessary to perform aspect identification on the review texts. Recently, Lu et al. \citep{r2} introduced a resource for aspect identification in peer review text. We obtained all data corresponding to the novelty aspect of reviews from this resource. We then evaluated both prior aspect identification models and LLMs on this data, with the specific results reported in Appendix \ref{novelty_eval}. Based on these results, we selected GPT-4o-mini as the model for automatic extraction of novelty evaluations. Using this model, we extracted all content related to novelty evaluation from reviewer reports of EMNLP 2023.

\subsubsection{Sentiment-Based Structuring of Novelty Evaluations}
During the review process, multiple reviewers may independently praise a paper for proposing a novel approach. To facilitate the evaluation of novelty evaluations generated by the LLM, we designed a prompt that instructs GPT-4o to remove redundant evaluations and organize novelty evaluations according to sentiment polarity. The specific details are provided in Appendix \ref{novelty_norm}.
\begin{table}[ht]
    \centering
    \small
    \begin{tabular}{lm{0.8cm}<{\centering}m{1.5cm}<{\centering}m{1.5cm}<{\centering}}
    \toprule
         Dataset& \#Paper& Avg. Nov Desc & Avg. Nov Eval\\
    \midrule
         COLING 2020& 87&6.1&-\\
         \rowcolor{orange!10}
         NovBench& 1,684&5.3&7.7\\
    \bottomrule
    \end{tabular}
    \caption{Dataset statistics of Novbench and COLING 2020 (subset in Figure \ref{overview}), including paper count, average novelty description (Avg. Nov Desc) sentence count, and average novelty evaluations (Avg. Nov Eval) count. '-' indicates not applicable.}
    \label{tab:1}
\end{table}
\begin{figure}[ht]
  \includegraphics[width=\columnwidth]{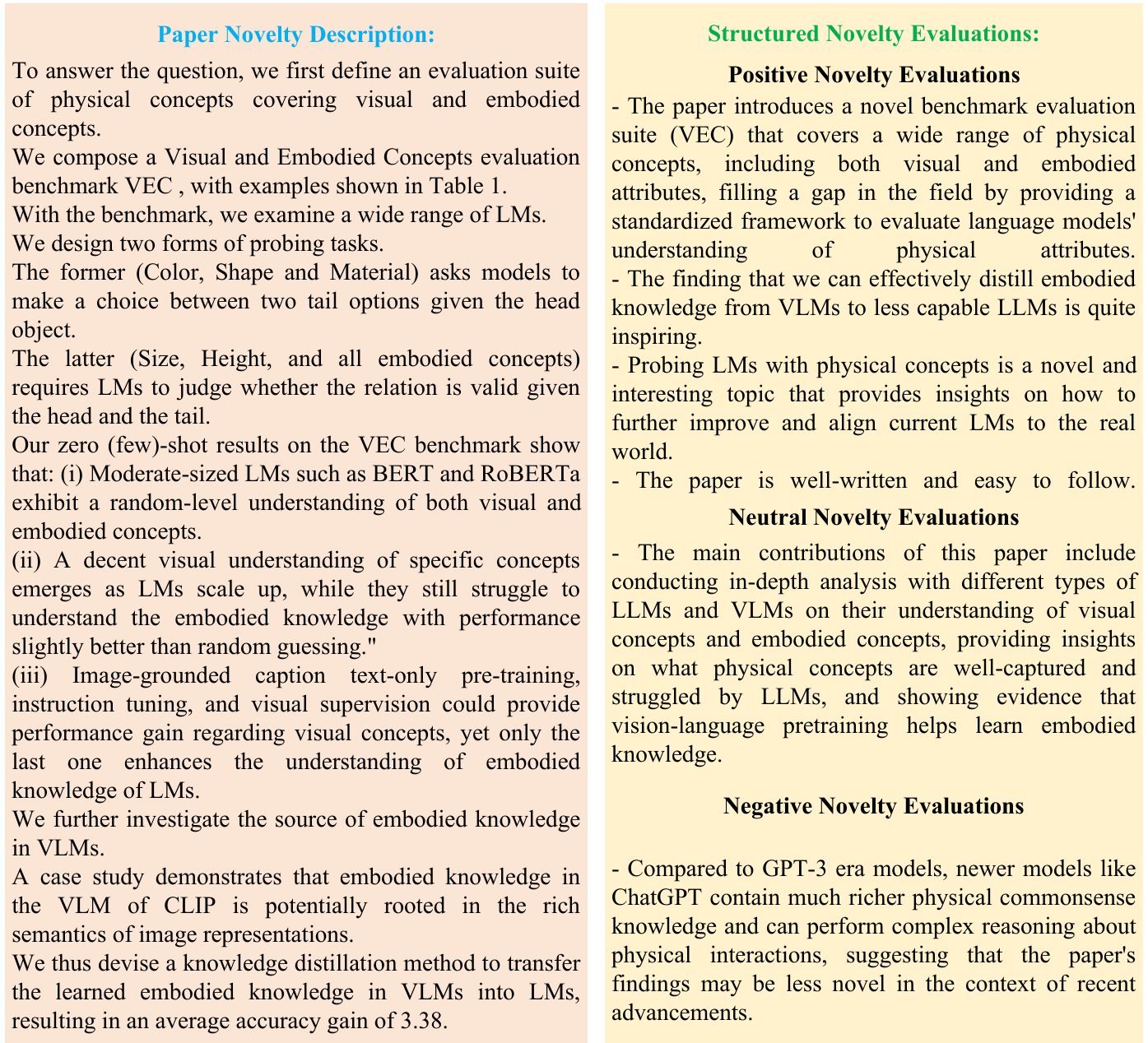}
  \caption{One Example of NovBench. The left side shows the novelty descriptions from the paper introductions, while the right side presents the structured novelty evaluations.}
  \label{fig:exp}
\end{figure}
\subsubsection{Dataset Statistics}
Table \ref{tab:1} summarizes the datasets used in this study. The COLING 2020 dataset corresponds to a manually annotated subset of NovBench. This subset includes only novelty descriptions from paper introductions and is primarily used to evaluate the performance of automatic novelty description extraction models, rather than for large-scale benchmarking. Such a design supports controlled model selection for the novelty description extraction stage. NovBench constitutes the full benchmark constructed in this work. For each paper, we automatically annotate two sources of novelty-related information: (1) novelty descriptions extracted from the paper introduction, and (2) novelty evaluations written by human reviewers. This dual-source design captures both author-stated novelty claims and independent human evaluations, enabling a systematic evaluation of LLM-generated novelty evaluations against human judgments. An example of the dataset is illustrated in Figure~\ref{fig:exp}.

\subsection{Dataset Evaluation Protocol}
In this work, we define the task as follows: given the novelty description from an academic paper's introduction, the LLM is required to generate an assessment of the novelty according to specific instructions. The output format mandates a structured output based on sentiment polarity.

Academic peer review is typically conducted along multiple quality dimensions rather than a single overall score. 
Reviewer guidelines\footnote{https://aclrollingreview.org/reviewerguidelines} for major NLP venues explicitly encourage assessments of a paper’s novelty, correctness or soundness, relevance or impact, and clarity of presentation. Such multi-dimensional evaluation practices motivate our design of four dimensions for assessing LLM-generated novelty evaluations. We fix the evaluation rubric to ensure a controlled and comparable setting, allowing us to isolate the novelty evaluation capability of LLMs.

\textbf{Relevance.} This dimension is defined as the degree to which the LLM-generated evaluation accurately comprehends the novelty description presented in the paper's introduction. To quantify this, we calculate the Information Matching Score (IMS) between the model-generated evaluation and the novelty description in the introduction, utilizing a method \citep{r3} designed to measure the information alignment between two scientific sentences. Our objective is to determine if B (LLM-generated evaluation) truly understands the content of A (the source text). Therefore, we adopt the \textit{Maximum Matching Average IMS} (AvgIMS), which measures how well each sentence in the source text is semantically covered by the most relevant sentence in the evaluation. For each sentence in $A$, we identify the review sentence that yields the highest IMS value, and compute the average of these maxima across all sentences:
\begin{equation}
        \textit{AvgIMS} = \frac{1}{N} \sum_{i=1}^{N} \max_{j} S_{ij}
\end{equation}
where $S_{ij}$ denotes the IMS between the $i$-th sentence in the source text and the $j$-th sentence in the LLM-generated evaluation. A higher AvgIMS indicates that the evaluation closely aligns with and accurately reflects the novelty description. 

\textbf{Correctness.} This dimension assesses the agreement between the model-generated positive, neutral, or negative novelty evaluations and human reviewers \citep{correctness}. To achieve this, we compare the resulting sentiment distribution produced by the LLM against the distribution established by the human reviewers. Following prior distribution-matching metrics, we define correctness as:
\begin{equation}
    \textit{DistAcc}= 1-\frac{\sum\left | p_i-t_i \right |  }{2} 
\end{equation}
where $p_i$ denotes the proportion of model-generated evaluations labeled with sentiment class $i$, and $t_i$ represents the corresponding human evaluations proportion. The numerator computes the $L_{1}$ distance between the two distributions, and the division by $2$ normalizes the maximum possible distance to $1$. A higher DistAcc value indicates better alignment between model generation and human sentiment judgments.\\
\textbf{Coverage.} To evaluate whether the LLM-generated novelty evaluations adequately capture the key points identified by human reviewers, we define a \textit{Coverage} dimension. Let $G$ denote the set of review novelty evaluations, and $M$ denote the set of LLM-generated evaluations. For each review novelty evaluation $g \in G$, we compute the cosine similarity with all LLM-generated evaluation $m \in M$, and count $g$ as covered if the maximum similarity exceeds a threshold $\tau$. Formally, Coverage is defined as:
\begin{equation} 
\textit{Coverage}= \frac{1}{|G|} \sum_{g \in G} \mathbf{1}\left[\max_{m \in M} \cos(g, m) \ge \tau \right],
\end{equation}
where $\cos(g, m)$ denotes the cosine similarity between the embeddings of $g$ and $m$, and $\mathbf{1}[\cdot]$ is the indicator function. In our experiments, we set $\tau = 0.7$, following common practice in prior work on sentence-level semantic similarity using sentence embeddings \citep{sentencebert}. A higher Coverage score indicates that more of the expert-identified novelty points are captured by the LLM-generated evaluations.\\
\textbf{Clarity.} The goal of this dimension is to determine if the generated review text is explicit and focused, ensuring the evaluation is easily understood and not overly generalized. We measure the clarity of LLM-generated novelty evaluations using a combined metric that accounts for both lexical grounding and sentence elaboration \citep{Clarity}. Let $K$ denote the set of keywords extracted from the introduction, and let $T$ denote the set of LLM-generated evaluation sentences. 

The first component, \textit{Keyword Coverage} (KC), assesses whether each generated sentence contains at least one introduction keyword, indicating lexical grounding in the source text:
\begin{equation}
\textit{KC} = \frac{1}{|T|} \sum_{t \in T} \mathbf{1}\Big[\exists k \in K \textit{ such that } k \subset t \Big]
\end{equation}

In our implementation, keywords are automatically derived from the introduction novelty sentences by extracting alphanumeric and hyphenated tokens using regular expressions, followed by filtering tokens with length greater than 5 to remove function words and generic short tokens. A generated evaluation is considered keyword-covered if it contains at least one such keyword (case-insensitive), and matching is performed at the token level to avoid spurious substring matches.
\begin{table*}[h]
\centering
\small
\resizebox{\textwidth}{!}{
\begin{tabular}{ccccc|cccc|cccc}
\toprule
\multirow{2}{*}{\centering\textbf{Model}} & \multicolumn{4}{c}{\textbf{Zero Shot}}& \multicolumn{4}{c}{\textbf{Few Shot}}& \multicolumn{4}{c}{\textbf{RAG}}\\
\cmidrule(lr){2-13}
&\textbf{Rel.} & \textbf{Cov.} & \textbf{Clarity} & \textbf{DistAcc} & \textbf{Rel.} & \textbf{Cov.} & \textbf{Clarity} & \textbf{DistAcc}& \textbf{Rel.} & \textbf{Cov.} & \textbf{Clarity} & \textbf{DistAcc}\\
\midrule
\rowcolor{blue!10}
\multicolumn{13}{c}{\textbf{General LLMs}}\\
\multicolumn{1}{l}{DeepSeek-R1-70B} &3.4885 & 0.2074 & 0.6470 & 0.6572 &3.4452 & 0.2112 & 0.6455 & 0.6274 & 3.0376 & 0.1500 & 0.6626 & 0.6260\\
\multicolumn{1}{l}{DeepSeek-R1-14B} & 3.4058 & 0.2053 & 0.6404 & 0.6436 &  3.3844 & 0.2252 & 0.6483 & 0.6606& 3.0715 & 0.1713 & 0.6628 & 0.6417\\
\multicolumn{1}{l}{DeepSeek-R1-8B} & 2.9475 & 0.1603 & 0.4949 & 0.5143 &  3.4703 & 0.2190 & 0.6160 & 0.6517 & 2.4988 & 0.1284 & 0.5459 & 0.5220\\
\multicolumn{1}{l}{Qwen3-32B} & 3.4747 & 0.2065 & 0.6609 & 0.6555 & 3.4175 & 0.2242 & 0.6497 & 0.6944 & 3.2555 & 0.1692 & 0.6679 & 0.6604\\
\multicolumn{1}{l}{Qwen3-14B} & 3.3892 & 0.1974 & \textbf{0.6634} & 0.6627 & 3.3656 & 0.2238 & 0.6487 & 0.6734 & 3.2015 & 0.1673 & 0.6681 & 0.6245\\
\multicolumn{1}{l}{Qwen3-8B} & 3.5769 & 0.1996 & 0.6471 & 0.6595 & 3.4061 & 0.2240 & 0.6487 & 0.6784 & 3.3042 & 0.1732 & 0.6593 & 0.6737\\
\multicolumn{1}{l}{GPT-4o} & \textbf{3.6983} & \textbf{0.2332} & 0.6595 & \textbf{0.6979} & \textbf{3.5609} & \textbf{0.2391} & 0.6587 & \textbf{0.7091} & 3.4481 & 0.2237 & 0.6668 & \textbf{0.6965}\\
\multicolumn{1}{l}{GPT-5} & 3.2772 & 0.1591 & 0.6209 & 0.4830 & 3.3124 & 0.1806 & 0.6164 & 0.5411 & 3.2300 & 0.1673 & 0.6666 & 0.6453\\
\multicolumn{1}{l}{gpt-oss-120b} & 3.2586 & 0.1787 & 0.6535 & 0.4376 & 3.1897 & 0.1830 & \textbf{0.6613} & 0.5107 & 3.1424 & 0.1644 & 0.6656 & 0.6027\\
\multicolumn{1}{l}{gpt-oss-20b} & 3.3158 & 0.1843 & 0.6270 & 0.5785 & 3.3298 & 0.2098 & 0.6364 & 0.6461 & 3.1372 & 0.1676 & 0.6269 & 0.6430\\
\multicolumn{1}{l}{Gemini-2.5-flash} & 3.4711 & 0.2120 & 0.6414 & 0.6011 & 3.4726 & 0.2364 & 0.6573 & 0.6590 & \textbf{3.5089} & \textbf{0.2270} & \textbf{0.6682} & 0.5923\\
\hline
\rowcolor{blue!10}
\multicolumn{13}{c}{\textbf{Specialized LLMs}}\\
\multicolumn{1}{l}{CycleReviewer-70B} &3.4632 & 0.2198 & 0.6598 & 0.6326 &3.3426 & 0.2209 & 0.6494 & 0.6522 & 3.0292 & 0.1522 & 0.6587 & 0.4892\\
\multicolumn{1}{l}{CycleReviewer-8B} & 3.0712 & 0.1577 & 0.6377 & 0.3837 & 3.0833 & 0.1892 & 0.6206 & 0.4088 & 2.8853 & 0.1336 & 0.5817 & 0.2785\\
\multicolumn{1}{l}{DeepReviewer-14B} & 2.7402 & 0.1173 & 0.6425 & 0.6356 & 2.7161 & 0.1140 & 0.6130 & 0.6301 & 2.7288 & 0.1193 & 0.6161 & 0.5246\\
\multicolumn{1}{l}{DeepReviewer-7B} & 2.4654 & 0.0745 & 0.5992 & 0.6134 & 2.6589 & 0.1051 & 0.6160 & 0.6308 & 2.5465 & 0.1004 & 0.5697 & 0.3907\\
\multicolumn{1}{l}{Llama-OpenReviewer-8B} & 2.1293 & 0.0604 & 0.3339 & 0.2216 & 2.3047 & 0.0823 & 0.5381 & 0.6205 & 1.2235 & 0.0317 & 0.2749 & 0.1031\\
\multicolumn{1}{l}{Reviewer2} & 1.8377 & 0.0408 & 0.4496 & 0.4517 & 0.9993 & 0.0013 & 0.3402 & 0.4083 & 0.1556 & 0.0000 & 0.0184 & 0.0709\\
\multicolumn{1}{l}{SEA-E} & 3.4259 & \underline{0.2610} & 0.6497 & 0.6834 & 3.3356 & \underline{0.2483} & 0.6395 & 0.6609 & 3.3807 & \underline{0.2712} & 0.6585 & 0.5965\\
\multicolumn{1}{l}{SEA-S} & \underline{3.6304} & 0.2576 & \underline{0.6630} & \underline{0.7162} & \underline{3.4091} & 0.2454 & \underline{0.6519} & \underline{0.7149} & \underline{3.5170} & 0.2474 & \underline{0.6662} & \underline{0.6740}\\
\midrule
\rowcolor{blue!10}
\multicolumn{1}{c}{Human} & 2.7899 & - & - & - & 2.7899 & - & - & - & 2.7899 & - & - & - \\
\bottomrule
\end{tabular}}
\caption{The evaluation performance of different models under various prompting strategies. For each metric, the best-performing general model is highlighted in bold, and the best-performing specialized model is \underline{underlined}. Rel. denotes Relevance, Cov. denotes Coverage, and DistAcc denotes Correctness. Human refers to the performance of human reviewers on our evaluation metrics. '-' indicates not applicable.}
\label{tab2}
\end{table*}
The second component, \textit{Length Score} (LS), encourages sufficiently informative sentences without enforcing verbosity by computing the average sentence length (in tokens), normalized by 20 and clipped to a maximum of 1:
\begin{equation}
\textit{LS} = \min\left(\frac{1}{20|T|} \sum_{t \in T} \textit{len}(t), 1\right)
\end{equation}
where $\textit{len}(t)$ denotes the token length of sentence $t$. Then,
to capture linguistic well-formedness and readability, we incorporate a fluency-based component derived from language model perplexity. Let $\textit{PPL}(t)$ denote the perplexity of sentence $t$ computed using a pretrained causal language model (distilgpt2 \citep{distilgpt2}). We define:
\begin{equation}
\textit{FS} = \frac{1}{|T|} \sum_{t \in T} \frac{1}{1 + \textit{PPL}(t)}
\end{equation}

Lower perplexity corresponds to higher fluency and better grammatical quality. The inverse transformation ensures that the score lies within $(0,1)$.
The final \textit{Clarity Score} is defined as the mean of these two components:
\begin{equation}
\textit{Clarity} = \frac{1}{3}(\textit{KC} + \textit{LS} + \textit{FS})
\end{equation}
A higher score indicates that the model generates sentences that are both lexically grounded and sufficiently elaborated.

Following the ACL/EMNLP review scoring system, Relevance is scored on a scale from 1 to 5, while the other dimensions are scored from 0 to 1.

\section{Experiments}
\subsection{Baselines Selection}
Based on our proposed evaluation metrics, we assessed a total of 11 general-purpose LLMs across two categories: \textbf{(1) Closed-source LLMs}: GPT-4o \citep{gpt4o}, GPT-5 \citep{gpt5}, and Gemini-2.5-flash \citep{gemini2.5}; (2) \textbf{Open-source LLMs}: DeepSeek-R1 (70B, 14B, 8B) \citep{deepseekr1}, Qwen3 (32B, 14B, 8B) \citep{qwen3}, and gpt-oss (120B, 20B) \citep{gptoss}. Furthermore, we also evaluated eight domain specialized LLMs that were \textbf{fine-tuned on peer review dataset}: CycleReviewer-70B, CycleReviewer-8B \citep{cycleresearcher}, DeepReviewer-14B, DeepReviewer-7B \citep{deepreviewer}, Llama-OpenReviewer-8B \citep{openreviewer}, Reviewer2 \citep{reviewer2}, SEA-E and SEA-S \citep{SEA}. We access closed-source models via their official APIs, while the open-source models were downloaded locally from HuggingFace\footnote{https://huggingface.co/} for inference. During testing on NovBench, we used greedy decoding with a maximum token limit of 4096 to guarantee output determinism and prevent truncation. We retained the default values for all other hyperparameters. We adopt three prompting strategies: zero-shot, few-shot, and Retrieval-Augmented Generation (RAG) \citep{rag}. Implementation details are shown in Appendix \ref{imple}. 
\subsection{Overall Performance of the Baseline Model with Automatic Metrics}
Table~\ref{tab2} reports model performance in different evaluation metrics and prompting strategies. From the results, we observe that across prompting settings, closed-source general LLMs (like GPT-4o and Gemini-2.5-Flash) have stronger performance, likely due to their larger parameter scales and undisclosed model architectures. When comparing models with comparable parameter sizes, specialized LLMs generally outperform general models, this advantage mainly depends on the choice of backbone and the fine-tuning strategy. For instance, SEA-S and SEA-E are built on Mistral (a mix-of-experts) backbone, which provides an inherent advantage for expert-level tasks such as novelty evaluation. Nevertheless, even with the same backbone, performance differences remain, driven by variations in fine-tuning approaches, as illustrated by CycleReviewer-8B and SEA-S. Overall performance tends to improve with increasing model size, though notable exceptions are observed for general-purpose LLMs. This may be because larger models’ stronger reasoning and generative abilities can induce over-interpretation and distributional drift under strict evaluation constraints. Additionally, from the results in the table, we observe that Human achieves relatively lower scores on the Relevance metric. This is because human reviewers typically rely on their domain knowledge and experience to make high-level judgments, rather than explicitly restating or strictly aligning their comments with the novelty descriptions in the paper introduction. 
\subsection{Human Agreement with Automatic Metrics}
To validate the effectiveness of our proposed metrics in assessing the generation of novelty evaluations, we randomly selected 100 samples for human evaluation. Specifically, we established a controlled comparison wherein evaluators were tasked with judging which model (Model A or Model B) produced the higher-quality novelty evaluation. This judgment was performed by providing the evaluators with the novelty description from the paper's introduction and the human reviewer's evaluation. The detailed examples and evaluation guidelines (Figure \ref{fig:guideline}) are provided in the Appendix \ref{agree}. Four human evaluators with strong expertise in Natural Language Processing, including two Ph.D. students, one Associate Professor, and one Lecturer, independently conducted the evaluation following the same guidelines. The inter-annotator agreement, measured by Fleiss’ $\kappa$, reached $0.72$, indicating substantial agreement. Our proposed automatic metrics demonstrated a high correlation with the corresponding human judgments (Spearman $\rho$ is $0.61$, with $p < 0.001$). This result confirms that our metrics are capable of correctly identifying superior model-generated evaluation, consistent with human preference (Agreement = $78\%$).
\section{Result Analysis}
\subsection{How does Different Prompt Strategies Affect Novelty Evaluation?}
For this question, we focus exclusively on the results pertaining to prompt tuning. As evidenced by the findings in Table \ref{tab2}, most models achieve their best performance in Relevance under the zero-shot prompting strategy. However, the maximum average score attained is only $3.6983$, which suggests that LLMs may unable to fully grasp the novelty described in the paper. Conversely, in the few-shot setting, model capability in Coverage and Correctness (DistAcc) demonstrates a noticeable improvement. However, this improvement is accompanied by a decrease in relevance. This trade-off suggests that when provided with human-evaluated examples, the LLM may be merely simulating human expression patterns and sentiment distribution rather than writing a genuine novelty evaluation. Furthermore, performance in clarity improves significantly in the RAG scenario. This outcome shows that the utilization of externally retrieved information helps in organizing and articulating the evaluation, resulting in output text with a clearer and more comprehensible structure. Simultaneously, the RAG approach leads to a reduction in relevance compared to both zero-Shot and few-Shot methods. This potential trade-off implies a key issue: while the retrieved information is comprehensive, the model may be mislead by the retrieval results in knowledge retrieval process. Consequently, this weakens its focus on the paper's novelty.

\subsection{Can Specialized LLMs Improve Novelty Evaluation?}
\label{sec5.2}
We hypothesized that models subjected to parameter fine-tuning on peer review datasets would exhibit better performance. However, the results presented in Table 2 indicate that these models only show a marginal advantages. Specifically, only the CycleReviewer-70B (large-parameter models), and the SEA series (contain data in NLP conferences), demonstrate better performance. We observe that CycleReviewer-70B and the SEA models maintain comparable scores in \textit{Relevance} while demonstrating superior performance over the general-purpose models across the other three dimensions. This finding suggests that while learning from human data results in a more anthropomorphic output style, it does not translate to a deeper, more robust understanding of novelty evaluation for this task. Furthermore, the Reviewer2 model performed particularly poor across all metrics. 
\begin{figure}[ht]
  \includegraphics[width=\columnwidth]{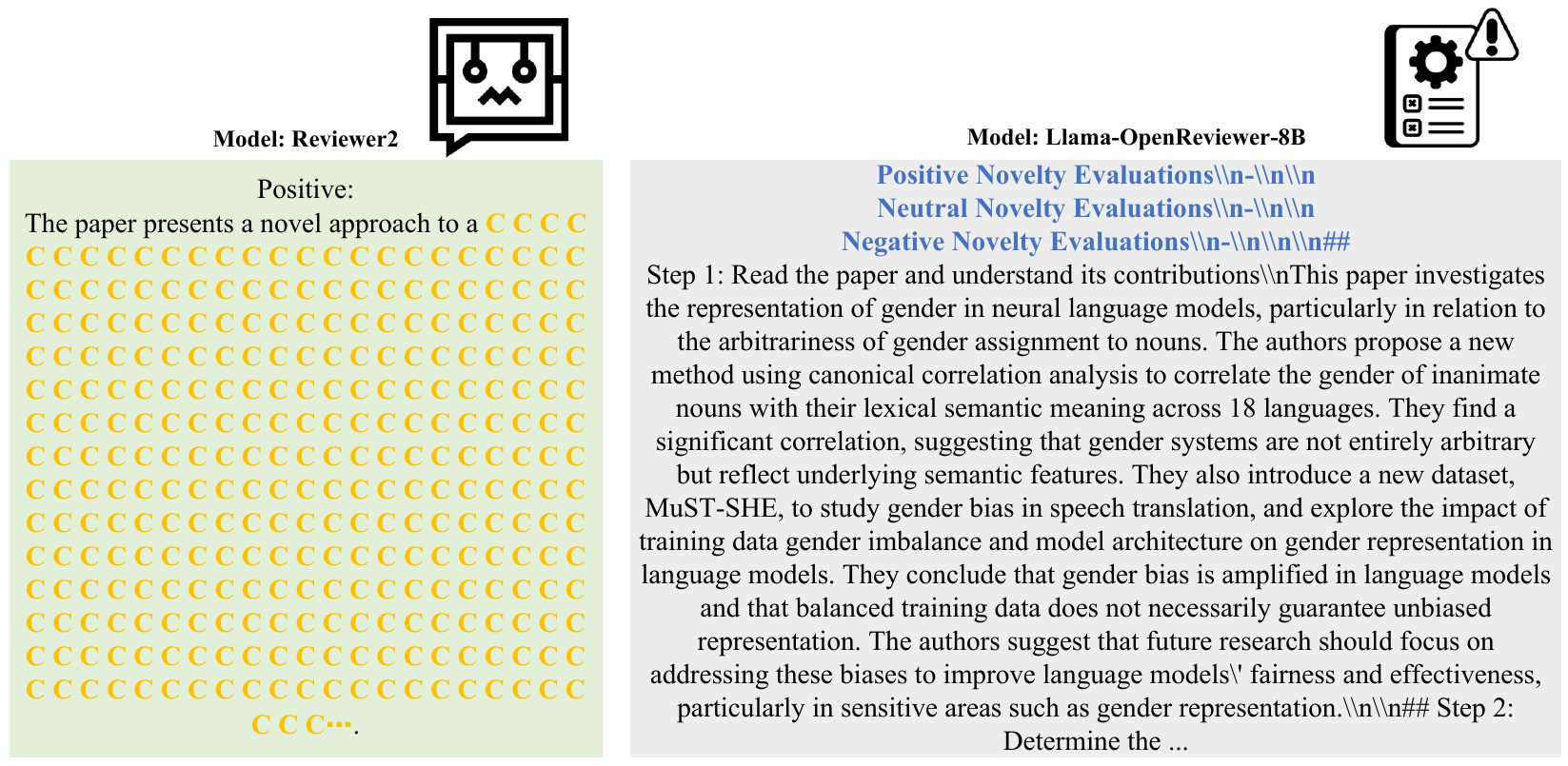}
  \caption{Examples of Instruction-Following Failures by the Specialized Model.}
  \label{fig2}
\end{figure}
An inspection of its generated output revealed a significant issue with instruction following, as illustrated in the accompanying Figure \ref{fig2}. We suspect that this model struggles to follow the given prompt instructions. This may be due to fine-tuning on highly specific training prompts, which weakens its general instruction-following ability. We checked that other specialized models (The detail in Appendix \ref{sec:instru}) exhibit similar problems, though the deficiency is most pronounced in Reviewer2. From these results, we think that models with a larger number of parameters and those designed to handle low-quality and inconsistent data are better equipped to provide strong instruction-following capabilities, rather than being fixed to a specific prompt.
\begin{figure}[ht]
  \includegraphics[width=\columnwidth]{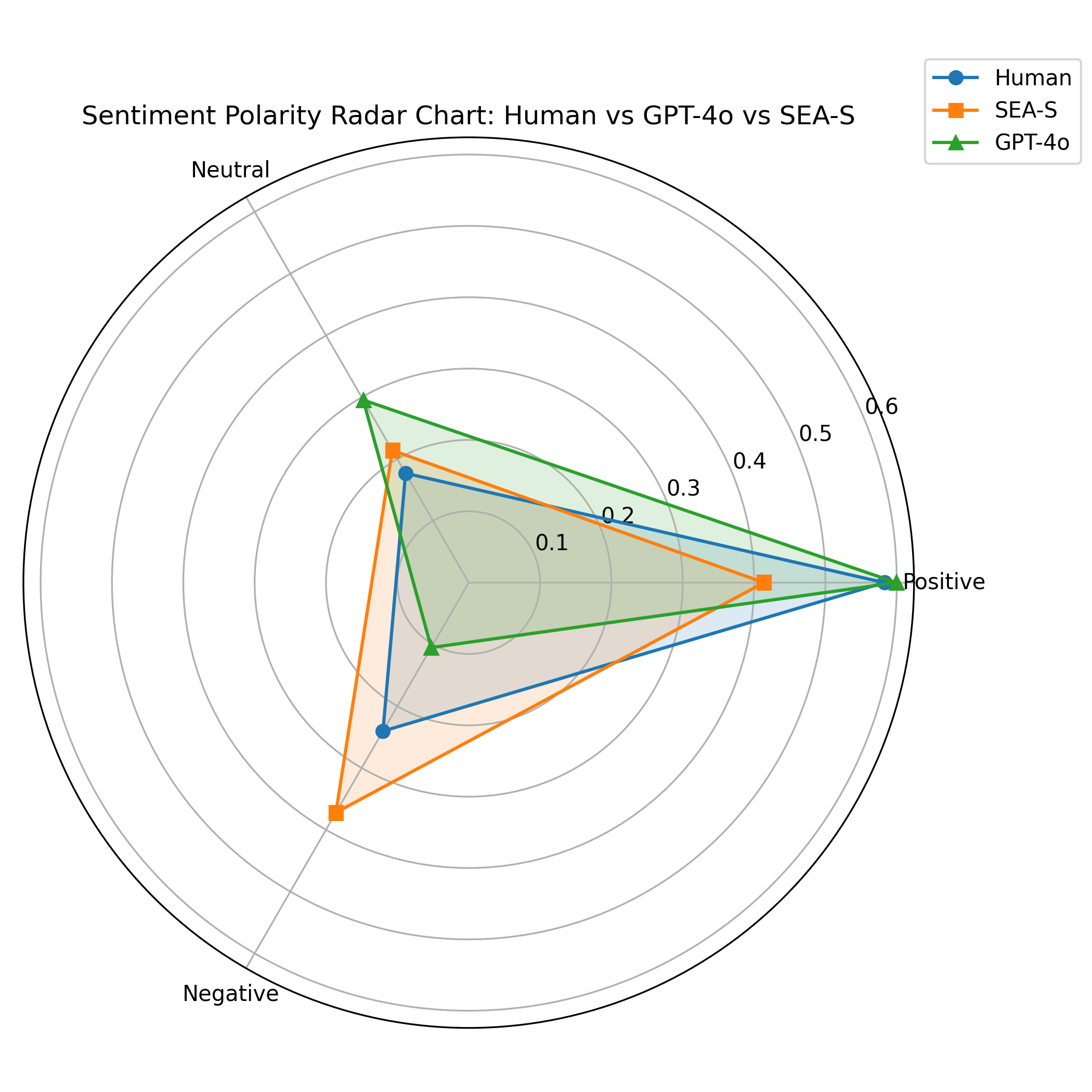}
  \caption{Comparison of Sentiment Polarity Distributions Among Human, General LLM, and Specialized LLM.}
  \label{fig3}
\end{figure}
\subsection{How Do LLM Novelty Evaluations Differ from Human Judgments?}
\label{sec5.3}
We selected two comparatively strong models, GPT-4o as a representative general model and SEA-S as a Specialized model, and evaluated their performance across all dimensions. As shown in the Appendix \ref{sec:comp}, for \textit{Relevance}, both models produce evaluations that are highly aligned with the novelty descriptions provided in the original papers. The generated positive evaluations in particular are almost entirely grounded in the explicitly stated methodologies and contributions. This suggests that both models are capable of identifying the core innovation claims of a paper. However, they also exhibit several issues, including exaggerating positive contributions, forcefully identifying negative aspects, introducing details not present in the source text, and producing overly templated and verbose assessments. For Coverage, the LLMs reliably capture the primary contributions, but they fall short in assessing the breadth of novelty. When a paper contains multiple innovation points, the models often fail to cover them comprehensively, potentially due to low sensitivity to different types of novelty. The models’ performance on Clarity is strong, indicating that they are able to extract and articulate the core concepts described in the paper. 
Finally, we compared the sentiment distributions of model-generated evaluations against human-written evaluations, as shown in the Figure \ref{fig3}. GPT-4o exhibits a distribution similar to humans for positive evaluations, but produces fewer negative evaluations and more neutral ones. In contrast, SEA-S displays the opposite trend: it produces substantially more negative and fewer positive evaluations. This suggests that general-purpose models tend to accommodate user expectations by generating more favorable feedback, whereas models fine-tuned on peer review data adopt a more critical stance, sometimes excessively so, potentially leading them to overemphasize or even fabricate negative points.
\subsection{Analysis of LLM Performance Across Novelty Evaluation Metrics}
Regarding \textit{Relevance}, although LLMs exhibit a surface-level understanding of novelty, they struggle to capture the specific and fine-grained content of novelty claims. This limitation is particularly evident under the RAG prompting setting, where performance degrades noticeably. These results indicate that retrieval augmentation or advanced prompting alone is insufficient to support genuine novelty understanding, and that specialized fine-tuning remains necessary.\\
\indent For \textit{Correctness}, better-performing specialized models achieve higher scores, suggesting that fine-tuning allows LLMs to learn human expressive and structural patterns. However, due to limited novelty understanding, these models often produce hedging evaluations with mixed sentiment, preventing optimal performance.\\
\indent Across all models, \textit{Coverage} remains sub-optimal. Even when restricted to novelty descriptions from the introduction, LLMs emphasize points that diverge from those identified by human reviewers. This highlights an important open challenge: enabling LLMs to better model how humans assess the breadth of novelty.\\
\indent In contrast, LLMs perform well on \textit{Clarity}, effectively identifying key term and major contributions in novelty descriptions, largely due to strong information extraction capabilities rather than a deeper understanding of novelty. \\
\indent Finally, we observe that some models fine-tuned on peer review data exhibit severe instruction-following issues, leading to substantial performance decrease and highlighting the need for improved fine-tuning strategies.
\subsection{Additional Analyses}
To further assess potential data contamination, temporal effects, and model behavior under different conditions, we conduct a series of additional analyses. Results show that model performance is largely consistent across model generations and publication years, and remains stable under controlled input perturbations, suggesting that performance is not driven by memorization or temporal leakage. 

We further analyze performance across paper types and reviewer disagreement. Models perform better on resource papers than methodological papers, indicating that evaluation difficulty varies with contribution type. Under reviewer disagreement, LLM-generated evaluations exhibit higher similarity to high-confidence reviews, suggesting non-arbitrary alignment behavior. 

Overall, these findings demonstrate that the proposed benchmark enables systematic and fine-grained analysis of LLM behavior beyond aggregate performance. Detailed results are provided in Appendix~\ref{sec:appendix_analysis}.
\section{Conclusion}
This paper proposes NovBench, a benchmark designed to systematically evaluate the ability of LLMs in assessing academic novelty. NovBench employs four distinct dimensions to quantify evaluation quality, utilizing a controlled and homogeneous setting to ensure reliability and isolate the novelty assessment task.  We demonstrated the performance of both general and specialized LLMs to evaluate academic paper novelty under varying prompting conditions. Through a comprehensive analysis of the novelty evaluations generated by different LLMs across all dimensions, we discuss key insights intended to guide future development in this field.  In future work, we plan to extend the benchmark to additional venues using the same data construction pipeline, enabling the study of cross venue and domain generalization of LLMs. Automatically deriving evaluation dimensions from reviewer guidelines is an interesting direction for future work, and our framework can be extended to support such dynamic rubrics.
\section*{Limitations}
This study is subject to several limitations. First, our work exclusively utilizes the paper introduction as the text source for novelty evaluation. While the introduction contains the primary novelty claims, relying solely on this section, rather than the full paper text, may omit detailed content required to fully support the evaluation.

Second, the data used are sourced from COLING and EMNLP proceedings, where the readily available peer review text predominantly corresponds to accepted papers, potentially introducing selection bias. In addition, our benchmark is constructed from a limited set of NLP venues, which may restrict its generalizability to broader research domains, as conferences such as ICLR and NeurIPS adopt different review formats, scoring rubrics, and cover broader interdisciplinary areas.

Third, we employ relatively simple prompt engineering strategies and do not explore more advanced prompting techniques or multi-agent architectures. Moreover, the credibility of reviewer comments remains an important concern, and we do not incorporate numerical scores (e.g., confidence scores) into the analysis.

Fourth, although EMNLP places greater emphasis on methodological novelty, our analysis does not distinguish between different types of novelty. Finally, despite the effectiveness of our proposed metrics, further research is needed to develop more robust evaluation methods.

Future work may explore more fine-grained taxonomy design, analyze hallucination patterns in novelty evaluation, and investigate multi-model aggregation approaches (e.g., ensembling or multi-agent methods) within the proposed framework. Despite these limitations, our study provides a useful reference for automated academic novelty assessment and LLM-based evaluation.

\section*{Ethics Statement}
This study is conducted in accordance with established ethical standards for research involving human-authored text. All data used in this work are openly available peer review reports released by conferences or journals, and do not contain personally identifiable information beyond what is already publicly disclosed. We do not collect new personal data, and our analysis poses no additional risk of privacy leakage or harm to authors or reviewers.

Importantly, the goal of this work is not to develop or promote automated peer review systems as a replacement for human expert reviewers. Instead, our focus is on evaluating the ability of large language models to assist in specific, well-scoped aspects of the review process—namely, the analysis and assessment of novelty—under controlled and transparent settings. We view such tools as potential supporting instruments that may help reduce reviewer workload or provide complementary perspectives, rather than substitutes for human judgment, expertise, or accountability.

We acknowledge the broader ethical concerns surrounding the use of LLMs in peer review, including risks of over-reliance, bias amplification, and misuse. Our work is intended to contribute empirical evidence that informs these discussions, rather than to advocate for the deployment of LLMs as autonomous reviewers.
\section*{Acknowledgments}
This work is supported by the Major Project of the National Social Science Fund of China (Grant No. 25\&ZD298). This research utilised Queen Mary's Apocrita HPC facility, supported by QMUL Research-IT \citep{qm}.

\bibliography{custom}

\appendix

\section{Supplement of Automatic Extraction of Novelty Descriptions}
\label{novelty_desc}

The accurate extraction of novelty descriptions from the paper introduction constitutes a critical step. We began by manually annotating the novelty descriptions within the introductions of the COLING 2020 papers sourced from NLPeer \citep{r1}, covering 87 papers, 2,300 total sentences, of which 533 were classified as novelty description sentences. The manual annotation process was executed by the two experienced journal and conference reviewers, we ask them to determine whether a given sentence constitutes a description of the paper’s novelty, with reference to the surrounding context, achieving a Cohen's $\kappa$ inter-rater agreement of $0.831$. We framed the automatic novelty description extraction as a binary classification task, where the model is required to identify whether a given sentence constitutes a novelty description. 
\begin{figure}[ht]
  \includegraphics[width=1.0\linewidth]{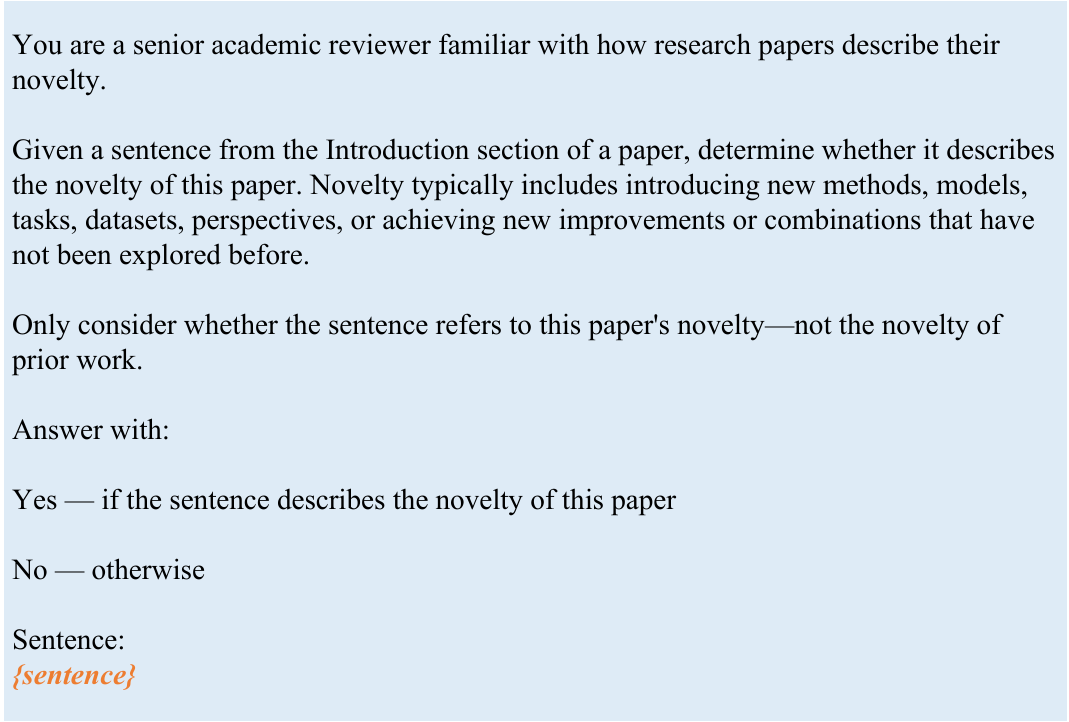}
  \caption{The Zero-Shot Prompt for Novelty Description Extraction.}
  \label{fig_zero_desc}
\end{figure}
\begin{figure*}[ht]
  \includegraphics[width=1.0\linewidth]{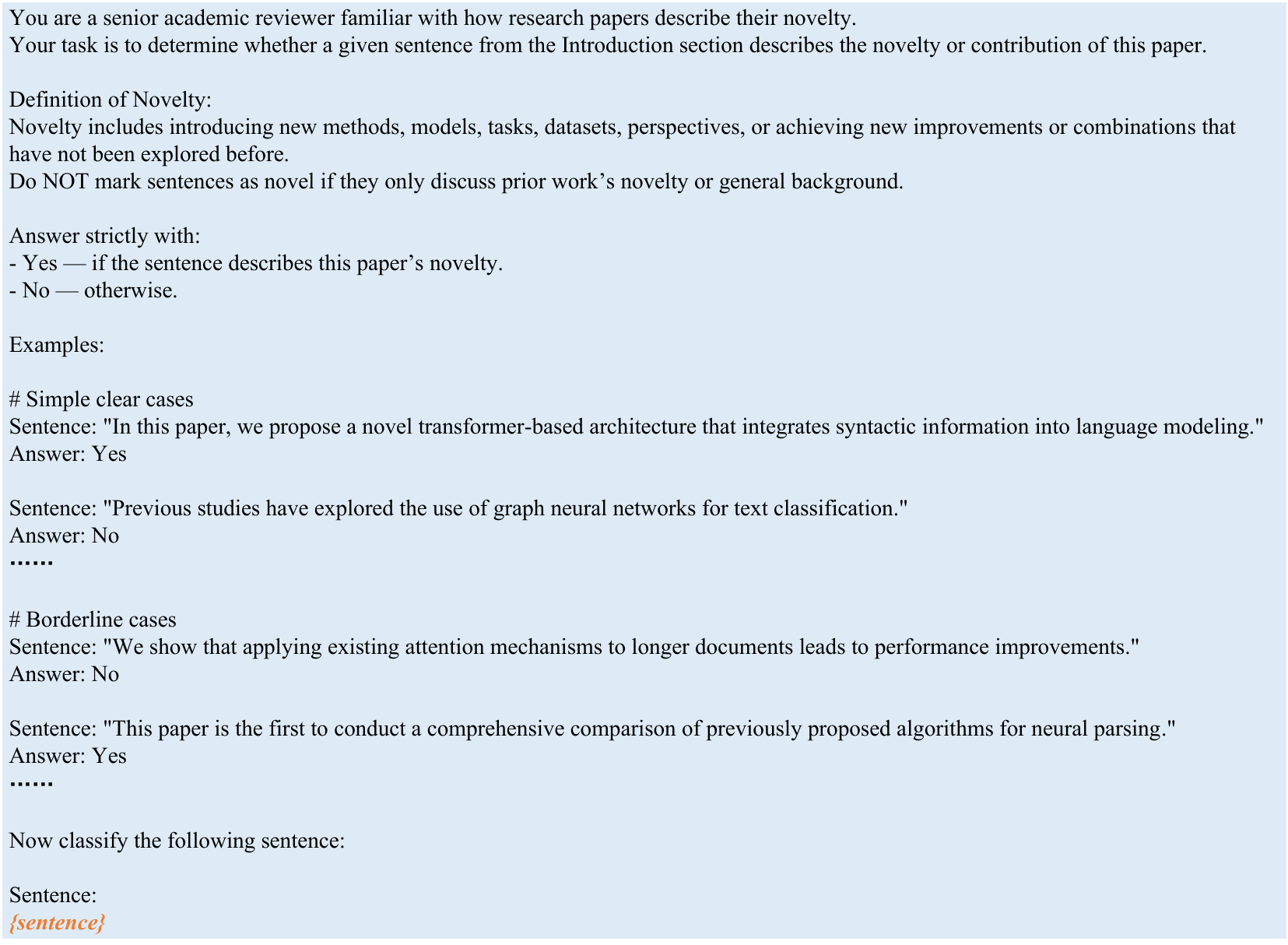}
  \caption{The Few-Shot Prompt for Novelty Description Extraction.}
  \label{fig_few_desc}
\end{figure*}
\begin{figure*}[ht]
  \includegraphics[width=1.0\linewidth]{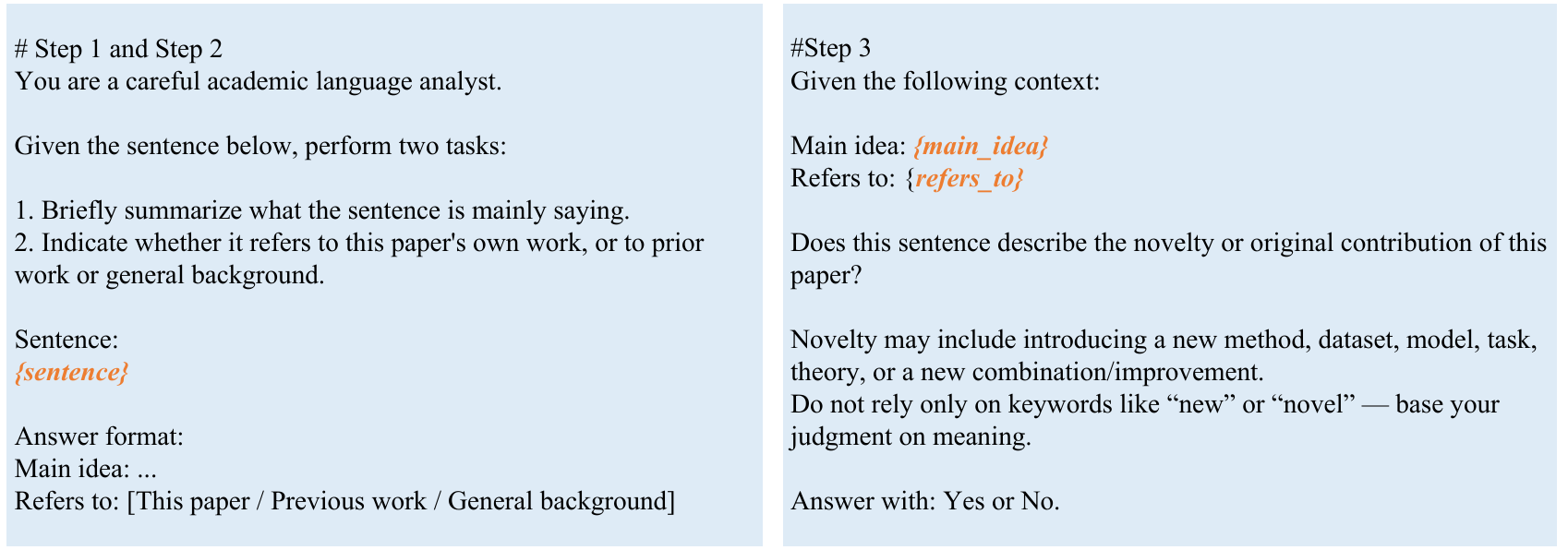}
  \caption{The Step by Step Prompt for Novelty Description Extraction.}
  \label{fig_step_desc}
\end{figure*}
\begin{figure*}[ht]
  \includegraphics[width=1.0\linewidth]{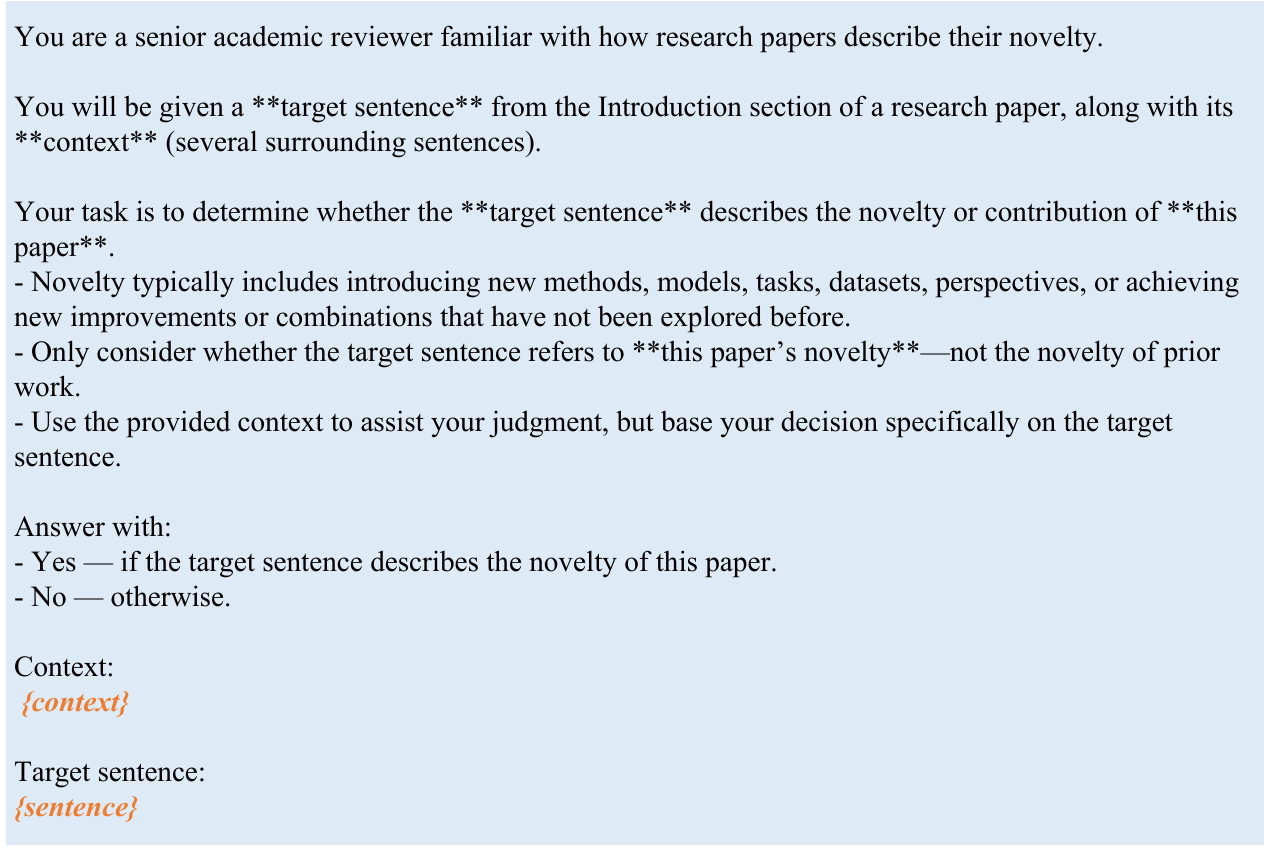}
  \caption{The Context Prompt for Novelty Description Extraction. We set the context window size to 2, meaning we utilized the two preceding sentences and the two succeeding sentences as contextual information. Boundary conditions were handled such that the first sentence included only succeeding context (post-text), and the last sentence included only preceding context (pre-text).}
  \label{fig_context_desc}
\end{figure*}
Specifically, we designed various prompting strategies (zero-shot see Figure \ref{fig_zero_desc}, few-shot see Figure \ref{fig_few_desc}, step-by-step see Figure \ref{fig_step_desc}, and in-context learning prompt see Figure \ref{fig_context_desc}) to benchmark the performance of various LLMs on this task. The results are presented in the Figure \ref{figs1}. 
\begin{figure*}[ht]
  \includegraphics[width=1.0\linewidth]{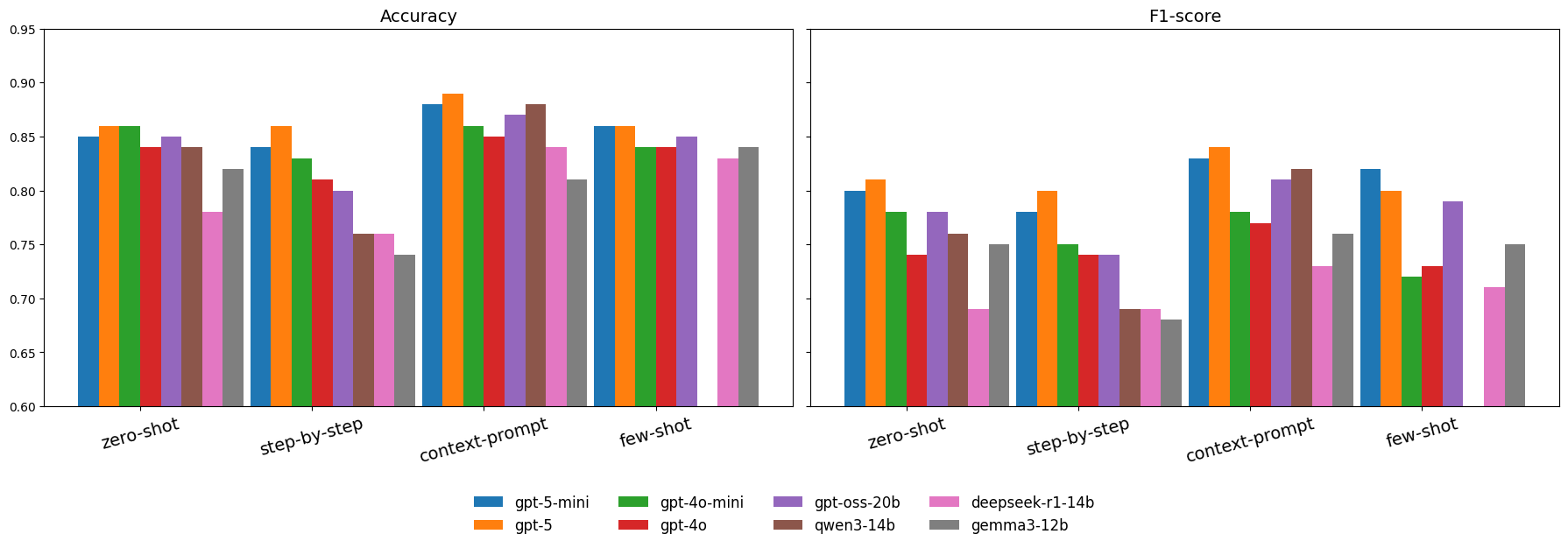}
  \caption{The performance of various LLMs on novelty description extraction under different prompt.}
  \label{figs1}
\end{figure*}
As shown in the Figure \ref{figs1}, the context prompt strategy yielded the best performance across all models, with GPT-5 achieving the highest Accuracy ($0.89$) and Macro F1 score ($0.84$). Consequently, we selected the context-prompted GPT-5 as the model for the automatic extraction of novelty descriptions. 
\begin{figure}[ht]
  \includegraphics[width=1.0\linewidth]{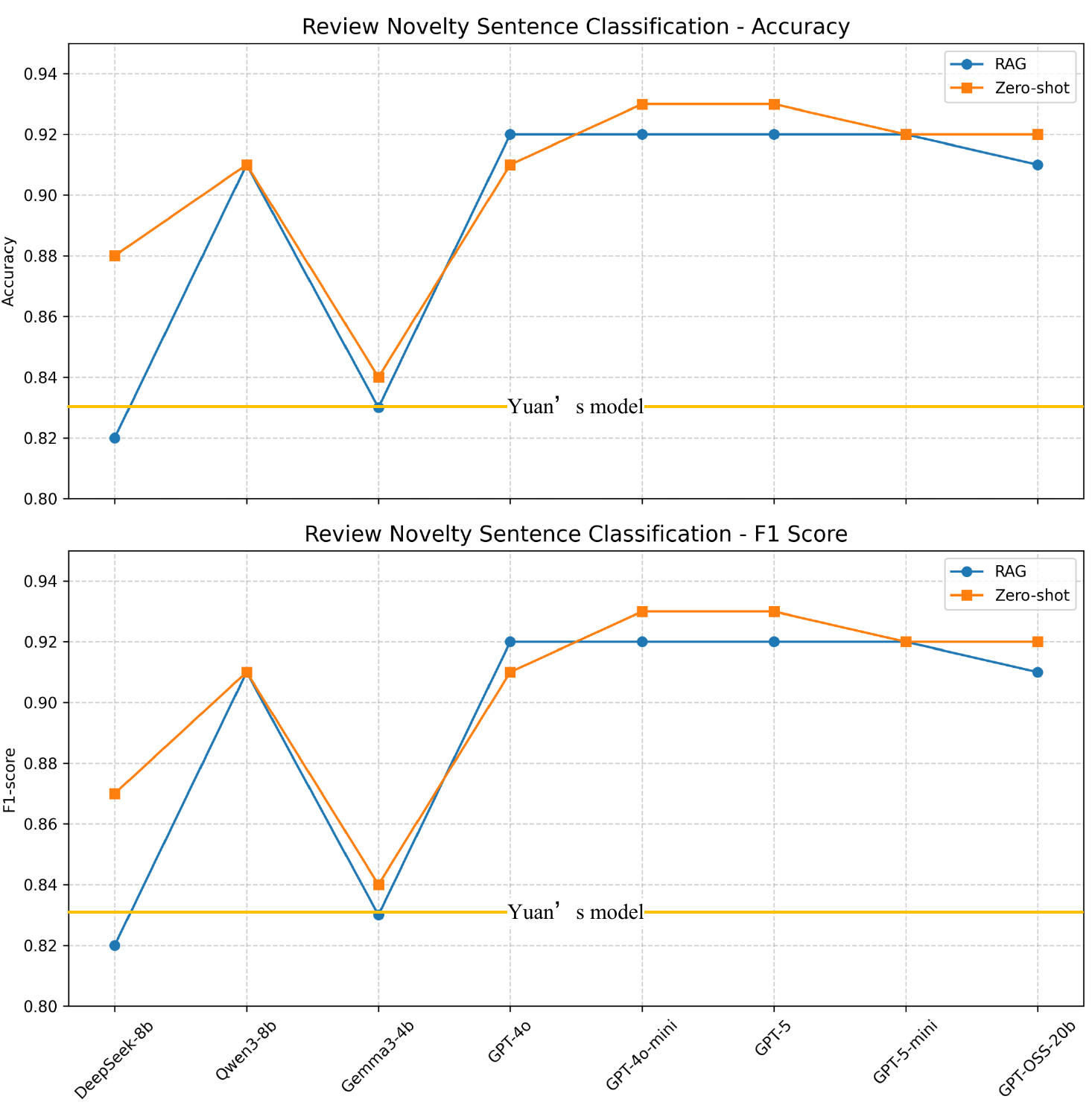}
  \caption{The performance of various LLMs on novelty evaluation extraction under different prompt.}
  \label{figs2}
\end{figure}
\begin{figure}[ht]
  \includegraphics[width=1.0\linewidth]{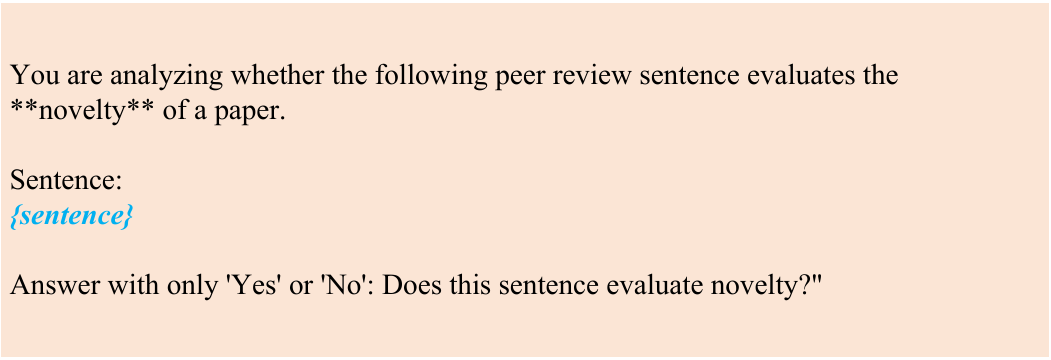}
  \caption{The Zero-Shot Prompt for Novelty Evaluations Extraction.}
  \label{fig_zero_eval}
\end{figure}
\begin{figure}[ht]
  \includegraphics[width=1.0\linewidth]{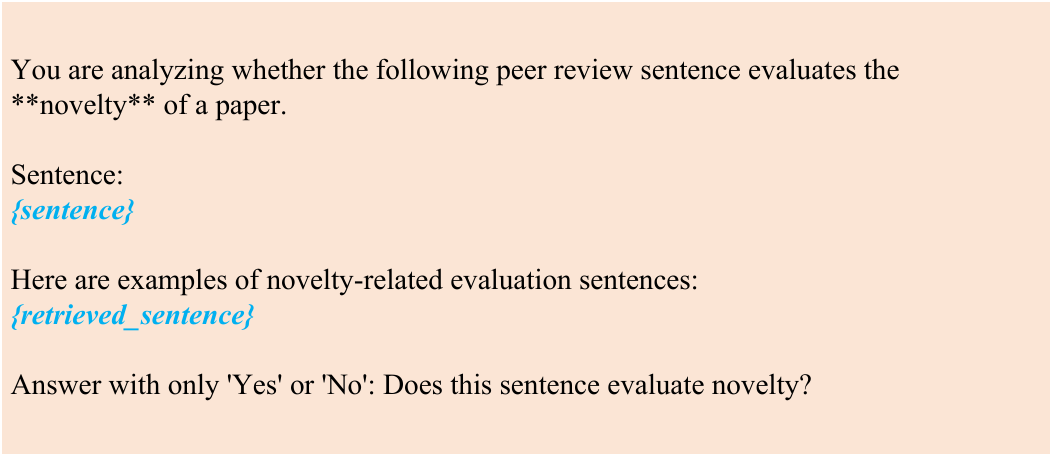}
  \caption{The RAG Prompt for Novelty Evaluations Extraction. The retrieved sentences were obtained by calculating the similarity between the query and the sentences related to novelty contained within the ReviewAdvisor \citep{rw4}.}
  \label{fig_rag_eval}
\end{figure}
\section{Supplement of Automatic Extraction of Novelty Evaluations}
\label{novelty_eval}
Similarly, the accurate extraction of novelty evaluations from the peer review text is equally crucial. We first obtained all novelty-related evaluation instances (totaling 493 comments) from the public resource shared by Lu et al. \citep{r2}, a dataset related to peer review aspect identification. We then randomly selected 500 instances of non-novelty evaluations, framing the task as a binary classification task for model testing. Specifically, given a review sentence extracted from the peer review text, the model is required to judge whether it is a novelty evaluation. We benchmarked the performance of the deep learning models provided by Yuan et al. \citep{rw4} against several LLMs, which executed the task under zero-shot (see Figure \ref{fig_zero_eval}) and RAG (see Figure \ref{fig_rag_eval}) prompt. The specific results are presented in the accompanying Figure \ref{figs2}. The results in the Figure \ref{figs2} indicate that GPT-4o-mini and GPT-5 achieved the best performance under the zero-shot prompting strategy, registering the highest combined Accuracy ($0.93$) and Macro F1 score ($0.93$). In consideration of cost-effectiveness, we designated the zero-shot prompted $\text{GPT-4o-mini}$ model as the model for extracting novelty evaluations.
\begin{figure*}[ht]
  \includegraphics[width=1.0\linewidth]{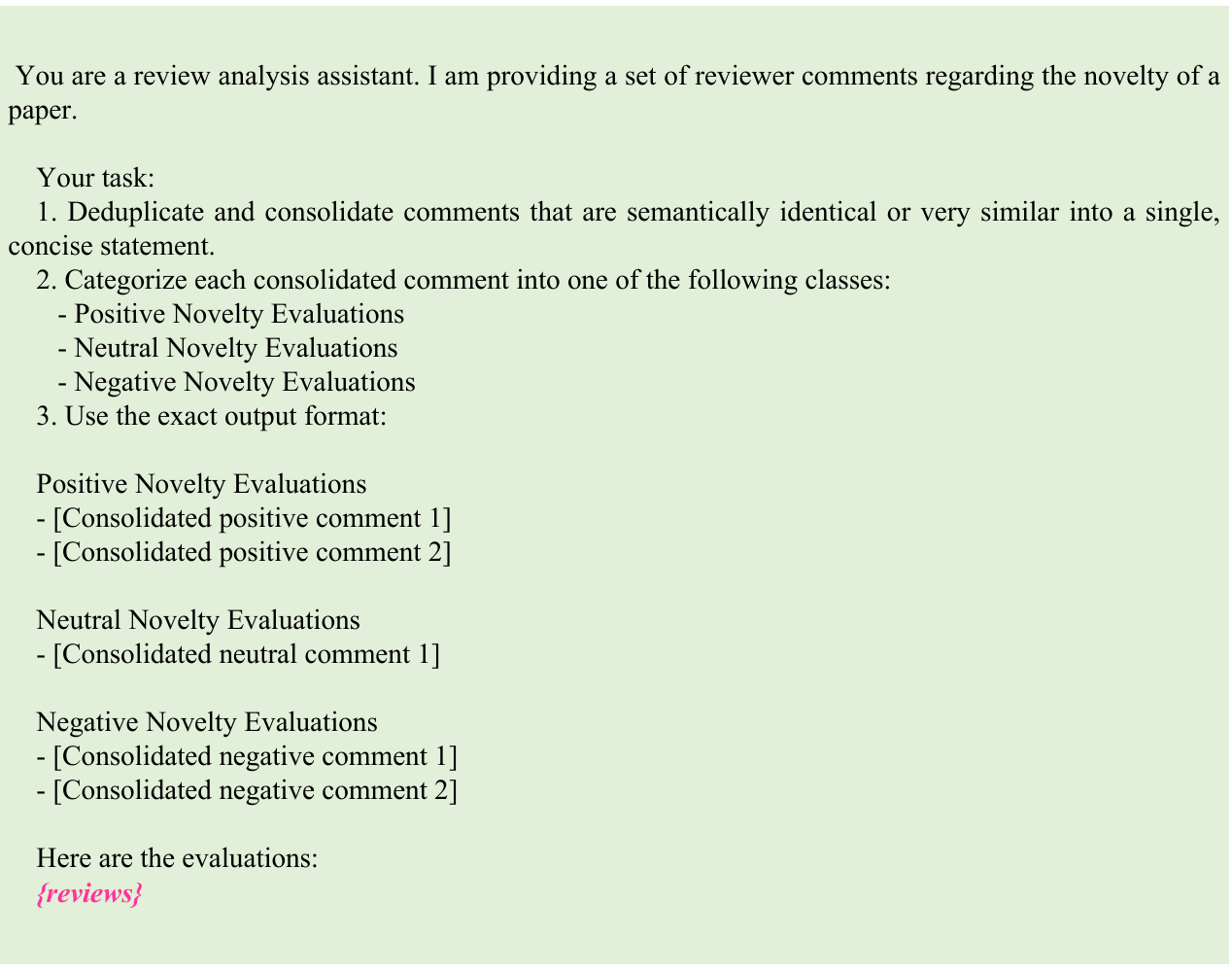}
  \caption{The Prompt for Structuring Novelty Evaluations based Sentiment.}
  \label{fig:prompt}
\end{figure*}
\section{Supplement of Sentiment-Based Normalization of Novelty Evaluations}
\label{novelty_norm}
To ensure fair comparison between human-written and LLM-generated evaluations, we use a prompt that instructs GPT-4o to (1) deduplicate semantically similar comments, (2) consolidate them into concise statements, and (3) categorize them by sentiment polarity. The exact prompt used in our experiments is reproduced below, and illustrated in Figure~\ref{fig:prompt}. This prompt ensures that all novelty-related feedback is standardized into a consistent and non-redundant set of evaluative statements, enabling more reliable automatic evaluation of novelty descriptions generated by LLMs.
\begin{figure*}[ht]
  \includegraphics[width=1.0\linewidth]{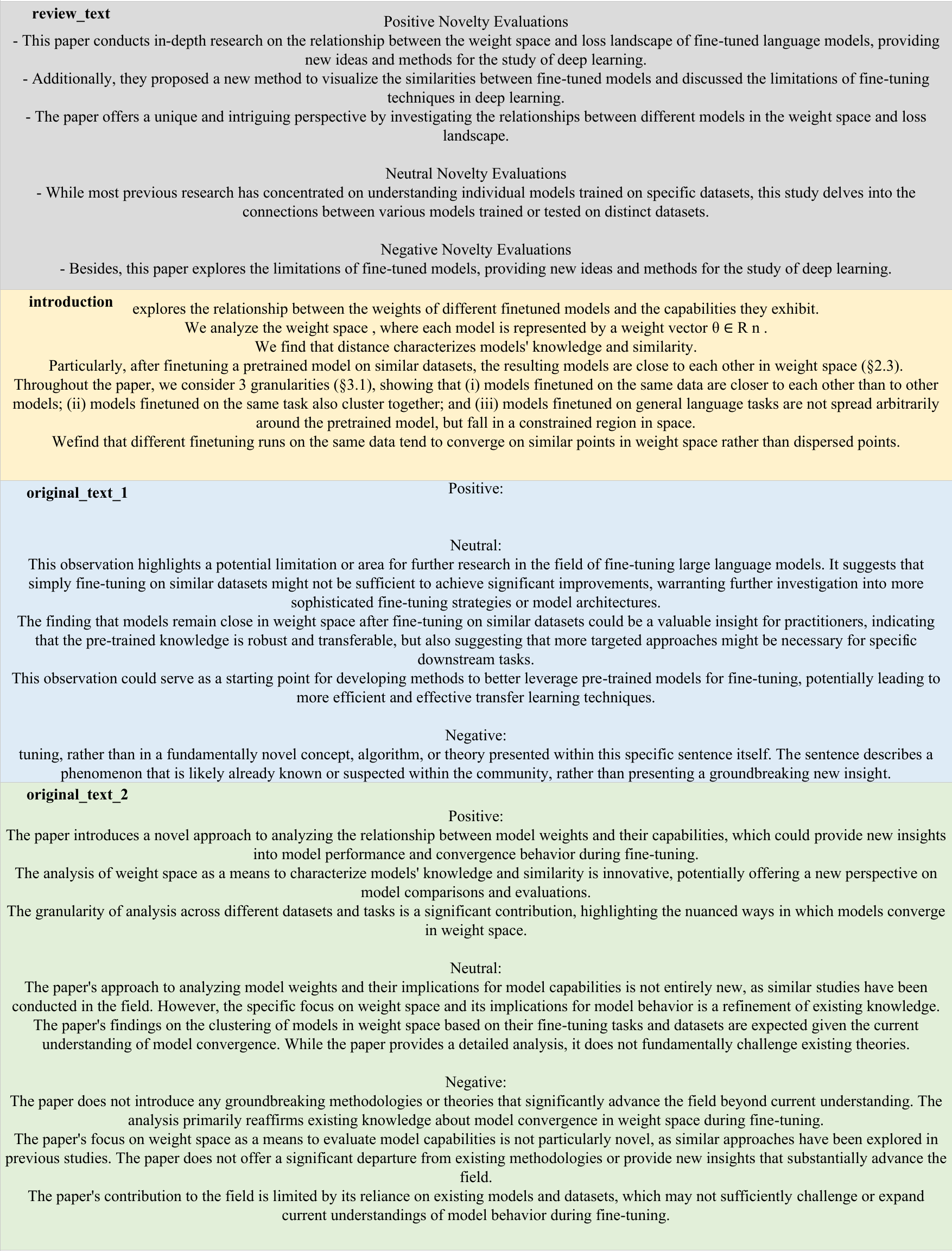}
  \caption{An Example for Human Evaluation.}
  \label{fig:human_ex}
\end{figure*}
\begin{figure*}[ht]
  \includegraphics[width=1.0\linewidth]{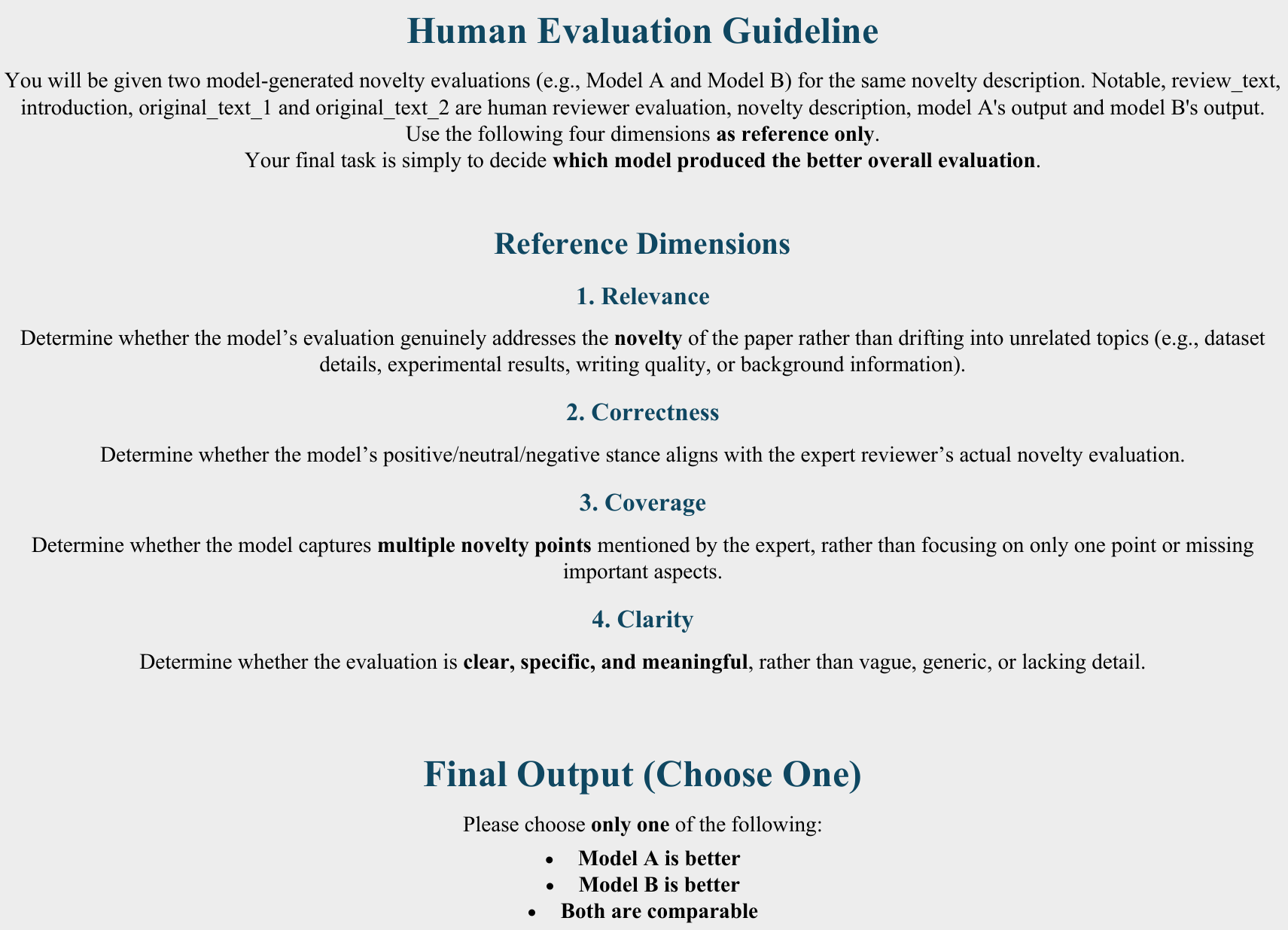}
  \caption{Guideline of Human Evaluation.}
  \label{fig:guideline}
\end{figure*}
\section{Supplement of Agreement Evaluation}
\label{agree}
This appendix provides the detailed instructions, examples (see Figure~\ref{fig:human_ex}), and guidelines (see Figure~\ref{fig:guideline}) used for the human evaluation of model-generated novelty assessments. We employ four human evaluators with strong expertise in Natural Language Processing (NLP), including two Ph.D. students, one Associate Professor, and one Lecturer. Each evaluator independently assesses, for each sample, which of the two models (Model A or Model B) produces a higher-quality novelty evaluation. The primary objective of this human evaluation is to validate the effectiveness of the proposed automatic evaluation metrics. Inter-annotator agreement is measured using Fleiss' $\kappa$ \citep{fleiss}, yielding a score of $0.72$, which indicates substantial agreement. To compare human judgments with automatic metrics, we compute both the Spearman rank correlation coefficient ($\rho$) and an agreement score that measures whether the metric selects the same preferred model as the aggregated human judgment. Formally, let $H^{(j)}_i$ denote the preference of the $j$-th annotator on sample $i$, where $j = 1, \dots, N$. The aggregated human preference $H_i$ is obtained via majority voting across annotators. Samples without a strict majority are excluded from the agreement computation. The agreement between the automatic metric and human judgments is defined as:
\begin{equation}
    \text{Agreement} = 
    \frac{1}{|\mathcal{D}|}
    \sum_{i \in \mathcal{D}} 
    \mathbf{1}(H_i = M_i),
\end{equation}
where $\mathcal{D}$ denotes the set of samples with valid aggregated labels, $M_i$ is the prediction of the automatic metric, and $\mathbf{1}(\cdot)$ is the indicator function.

\begin{figure*}[ht]
  \includegraphics[width=1.0\linewidth]{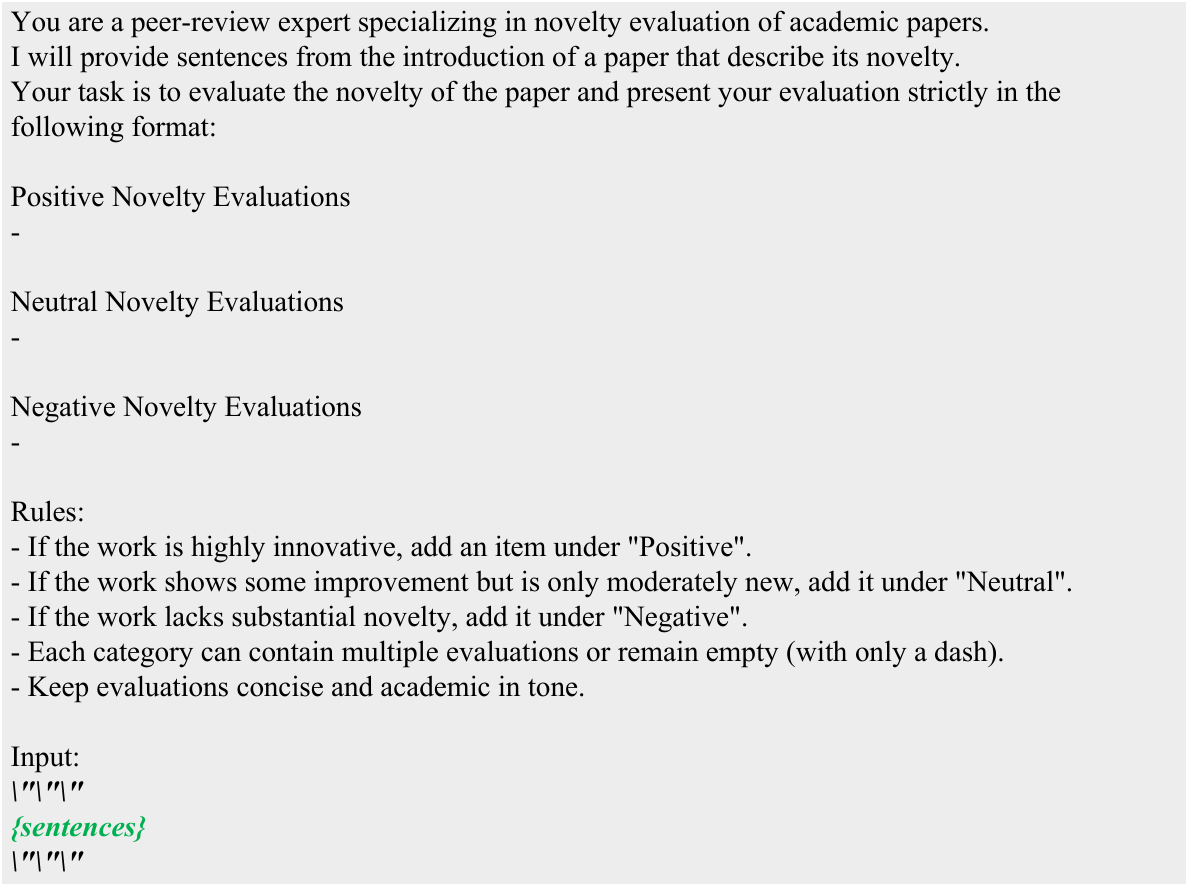}
  \caption{The zero shot prompt for different LLMs on NovBench.}
  \label{zero}
\end{figure*}
\begin{figure*}[ht]
  \includegraphics[width=1.0\linewidth]{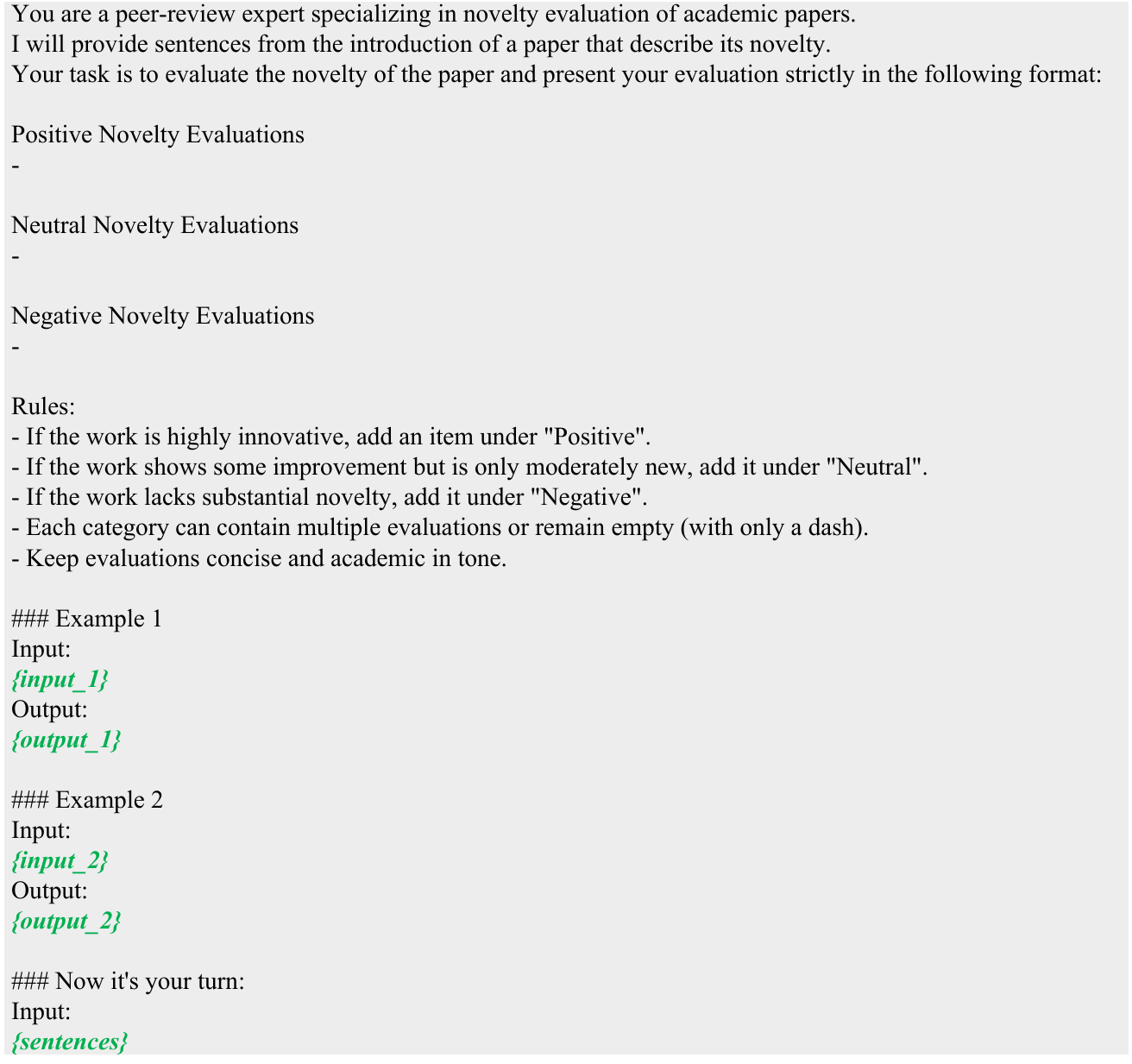}
  \caption{The few shot prompt for different LLMs on NovBench.}
  \label{few}
\end{figure*}
\begin{figure*}[ht]
  \includegraphics[width=1.0\linewidth]{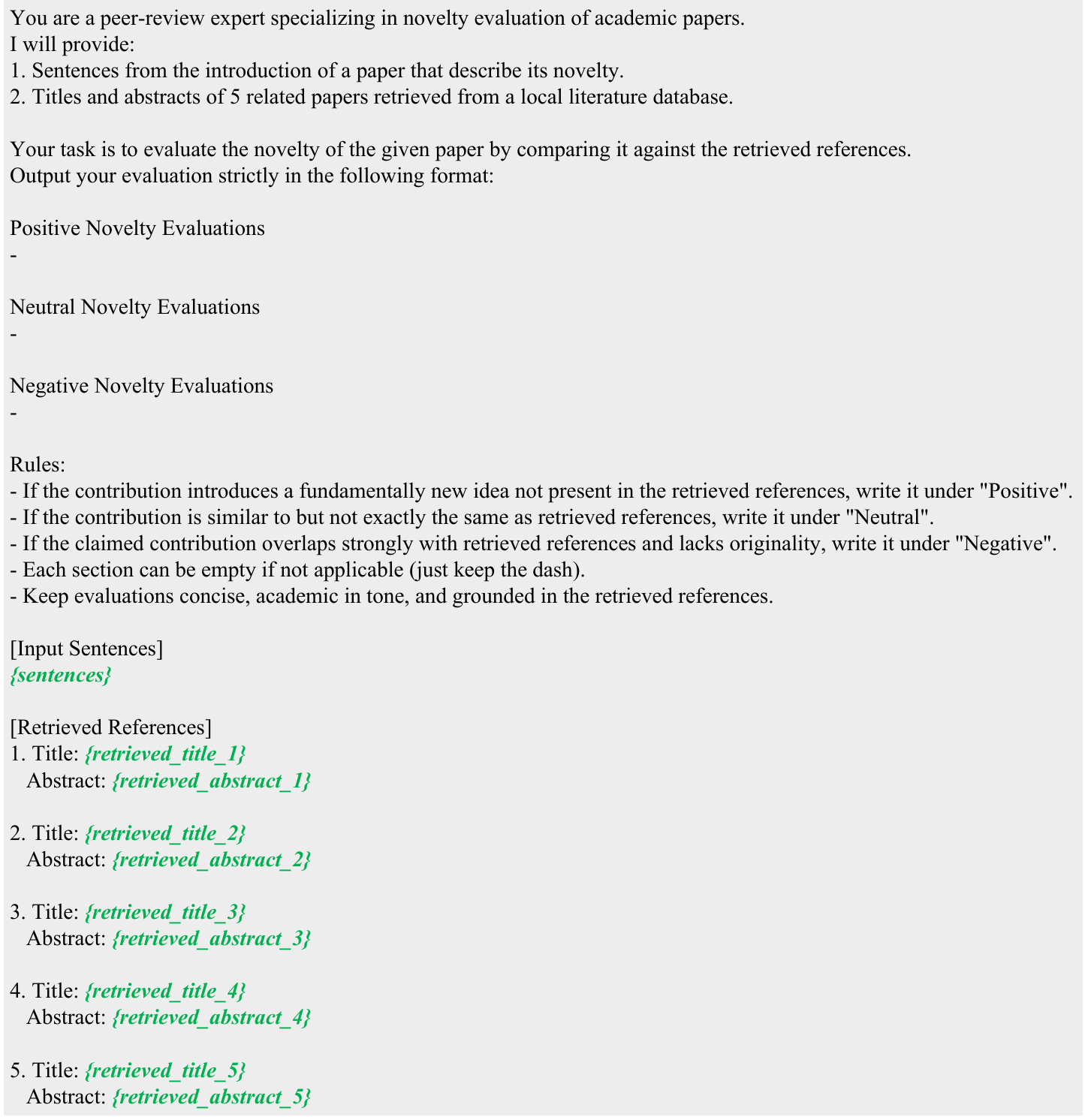}
  \caption{The RAG prompt for different LLMs on NovBench.}
  \label{rag}
\end{figure*}
\section{Experiment Implementation Details}
\label{imple}
During testing on NovBench, we evaluated various general and specialized LLMs using three distinct prompt tuning strategies: zero-shot (see Figure \ref{zero}), few-shot (see Figure \ref{few}), and RAG (see Figure \ref{rag}). For the zero-shot setting, the model is provided only with the extracted novelty descriptions. For the few-shot setting, the model receives the extracted novelty descriptions along with two analogous examples selected from our dataset as demonstrations. For the RAG setting, the model is provided with the extracted novelty descriptions and additional retrieved context, where the retrieval corpus consists of titles and abstracts of ACL, EMNLP, and NAACL papers published between 2019 and 2022, sourced from the ACL Anthology. Specifically, we utilized acl-anthology-helper (\url{https://github.com/tangg555/acl-anthology-helper}) to acquire and store the ACL Anthology papers in a local database. We then filtered this repository to include only the titles and abstracts from the specified ACL, EMNLP, and NAACL proceedings (2019–2022). Retrieval was executed using the abstract of each paper in NovBench as the query, ultimately yielding the 5 most relevant titles and abstracts per paper to serve as the RAG content.

Here, we provide additional details on the eight fine-tuned LLMs. The CycleReviewer (8B's backbone is Mistral-Nemo-12B\footnote{https://mistral.ai/news/mistral-nemo}, 70B's backbone is Qwen2.5-Instruct-72B \citep{qwen2.5}) series models are primarily fine-tuned on peer review data from ICLR 2024, covering the fields of machine learning and artificial intelligence. The DeepReviewer (backbone is Phi-4 \citep{phi4}) series models are mainly fine-tuned on peer review data from ICLR 2024 and ICLR 2025, also spanning machine learning and artificial intelligence. Llama-OpenReview-8B (backbone is Llama-3.1-8B-Instruct \citep{llama3}) is fine-tuned on peer review data from ICLR and NeurIPS (post-2022), covering machine learning and artificial intelligence. Reviewer2 (backbone is Llama-2-7B-Chat \citep{llama2}) is primarily fine-tuned on peer review data from NLPeer (CoNLL-16, ACL-17, COLING-20, ARR-22), ICLR 2017–2023, and NeurIPS 2016–2022, covering machine learning, natural language processing, computational linguistics, and artificial intelligence, with approximately 7B parameters. SEA-E and SEA-S are mainly fine-tuned on peer review data from NLPeer (CoNLL-16, ACL-17, COLING-20, ARR-22), NeurIPS 2016–2023, and ICLR 2017–2024, covering machine learning, natural language processing, computational linguistics, and artificial intelligence, and both backbone is Mistral-7B-Instruct-v0.2 \citep{mistral7b}.\\
\indent LLM inference was executed utilizing A100 80GB GPUs and H100 80GB GPUs. Specifically, models sized 8B, 14B, 20B, and 32B, along with CycleReviewer-8B, DeepReviewer-7B, Llama-OpenReviewer-8B, Reviewer2, SEA-E, and SEA-S, were run on a single A100 80GB GPU. Models at the 70B parameter scale and DeepReviewer-14B required inference to be distributed across two A100 80GB GPUs. Finally, the gpt-oss-120B model was allocated across two H100 80GB GPUs. It is important to note that we employed the Fast Mode configuration for all inferences involving CycleReviewer and DeepReviewer. The total inference time per model, contingent upon its parameter size, ranged from 5 to 70 hours. For Closed-source models, the inference process was implemented through official API integration..
\begin{figure*}[ht]
  \includegraphics[width=1.0\linewidth]{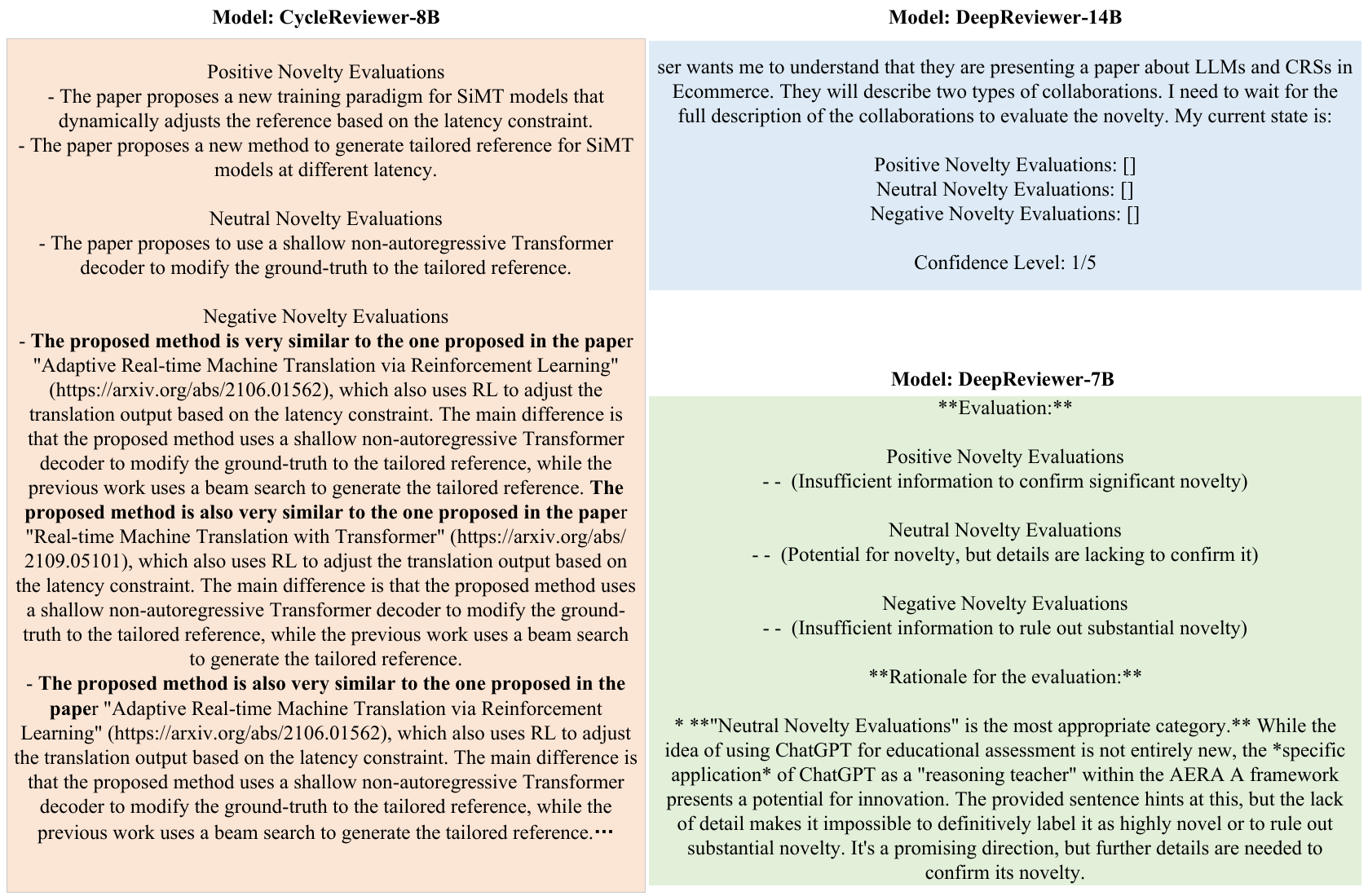}
  \caption{Examples of Instruction-Following Failures by other Specialized Models.}
  \label{fig:instru}
\end{figure*}
\section{Supplemental Analysis of Instruction-Following Deficiencies in Specialized Review Generation Models}
\label{sec:instru}
Beyond the particularly severe instruction-following deficiencies reported in Section \ref{sec5.2}, we observed that other models subjected to parameter fine-tuning on peer review feedback exhibit similar, and arguably unacceptable, operational failures. These specific issues are visually documented in the accompanying Figure \ref{fig:instru}. As depicted, CycleReviewer-8b suffers from the propensity to generate repetitive evaluations, whereas DeepReviewer exhibits the problem of producing null or empty evaluations.
\begin{figure*}[ht]
  \includegraphics[width=1.0\linewidth]{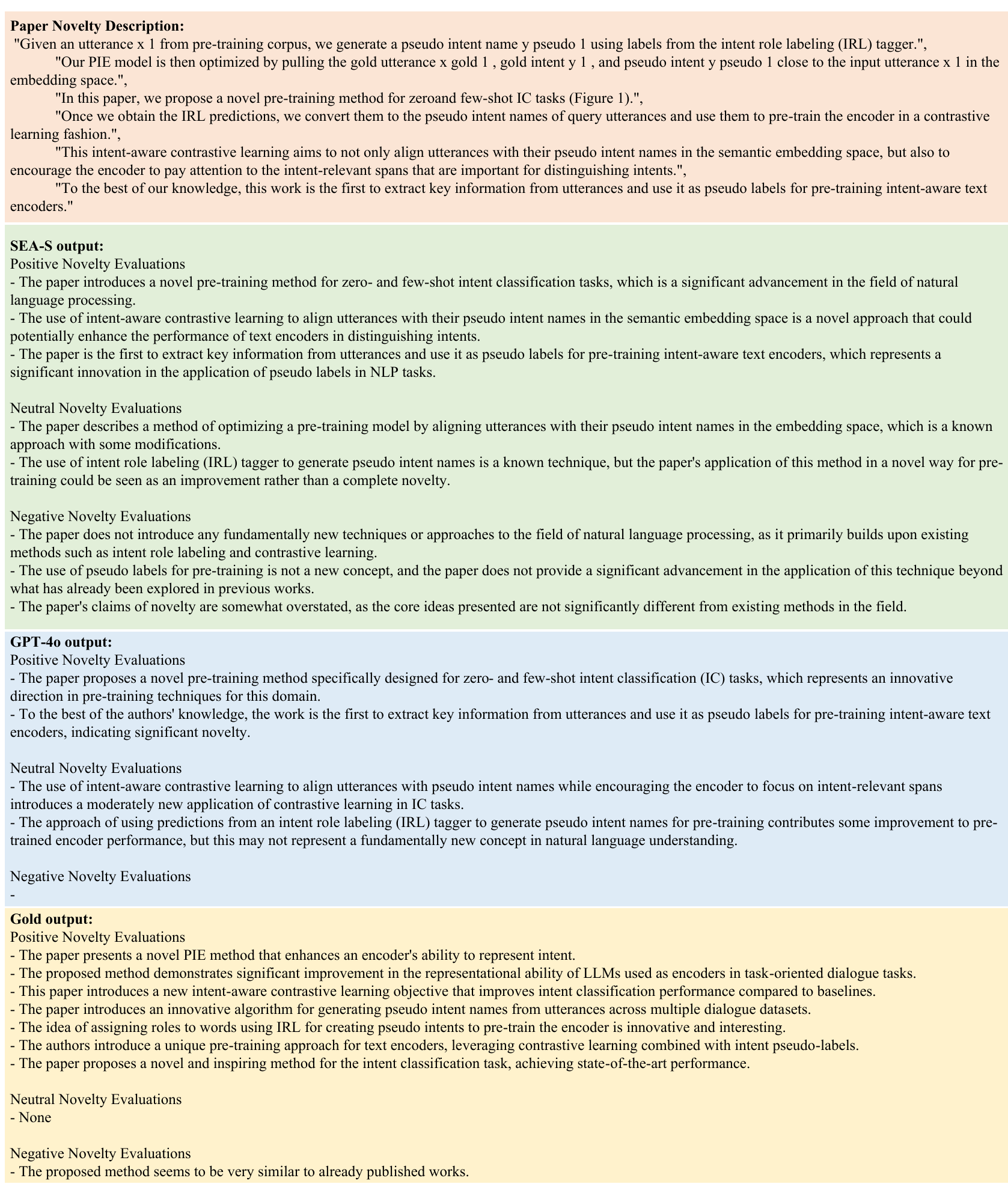}
  \caption{Case Outputs of $\text{SEA-S}$ and $\text{GPT-4o}$ Compared with Novelty Descriptions from the Paper Introduction and Human Reviewer Evaluations.}
  \label{fig:sam1}
\end{figure*}
\begin{figure*}[ht]
  \includegraphics[width=1.0\linewidth]{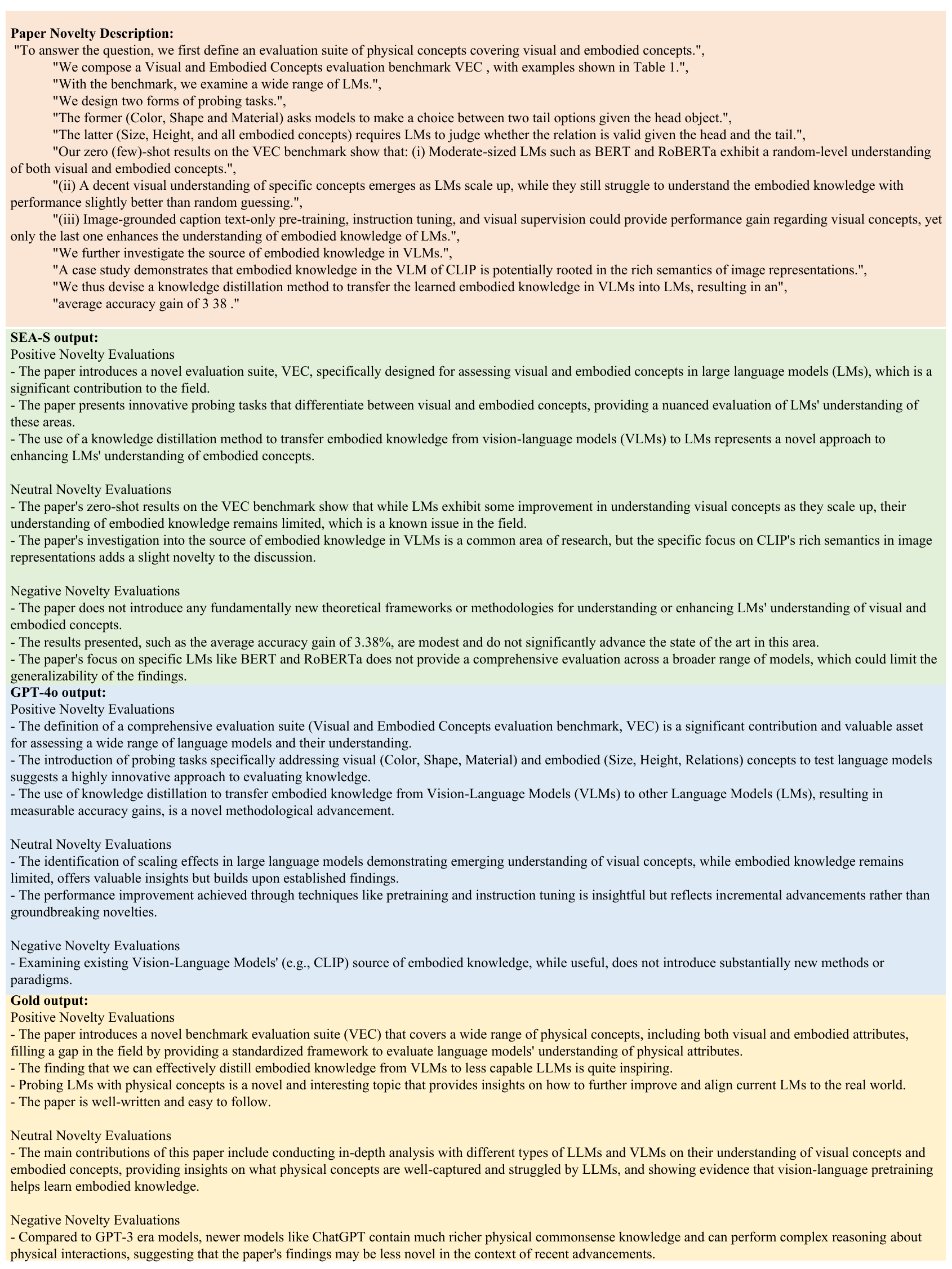}
  \caption{Case Outputs of $\text{SEA-S}$ and $\text{GPT-4o}$ Compared with Novelty Descriptions from the Paper Introduction and Human Reviewer Evaluations.}
  \label{fig:sam2}
\end{figure*}
\begin{figure*}[ht]
  \includegraphics[width=1.0\linewidth]{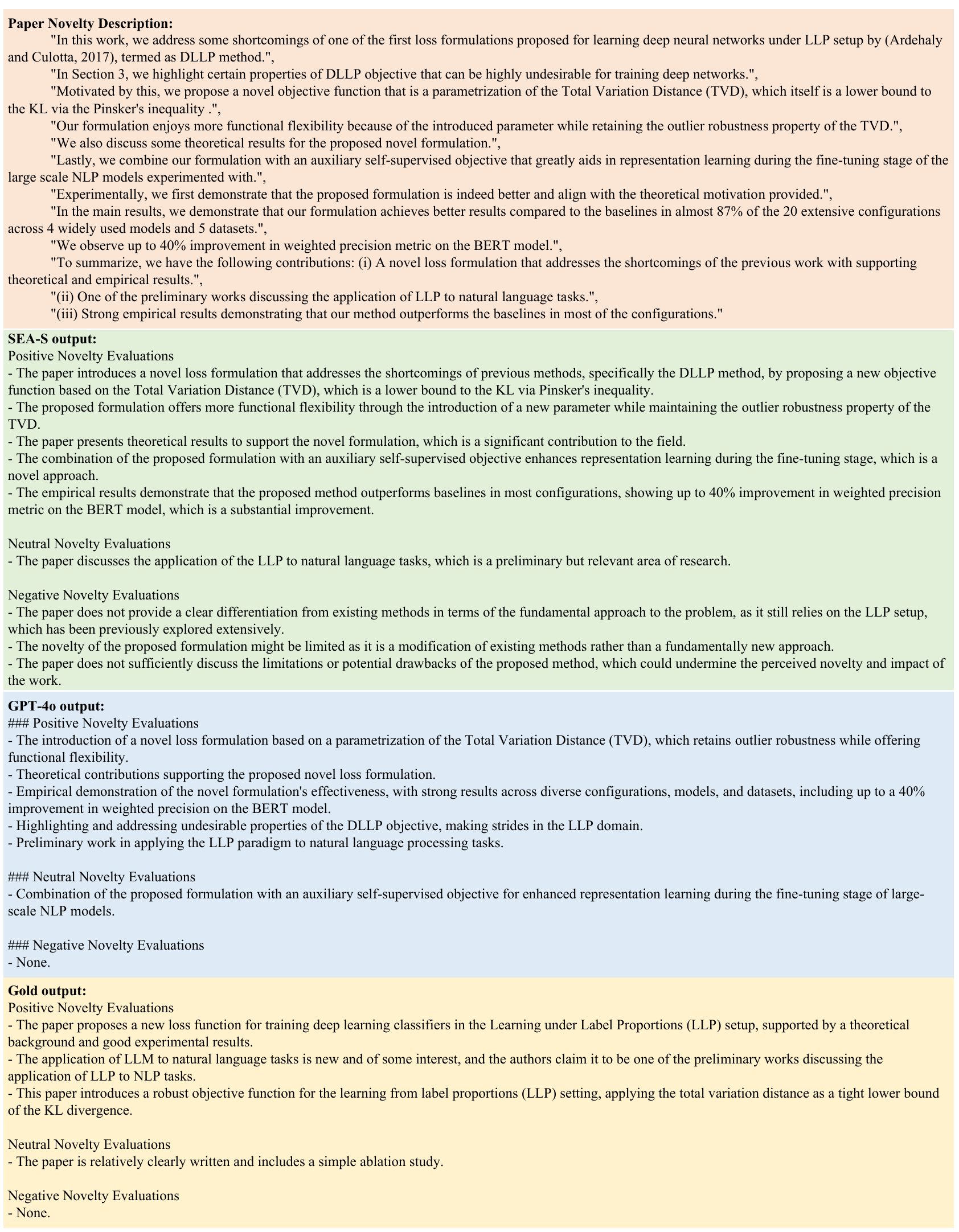}
  \caption{Case Outputs of $\text{SEA-S}$ and $\text{GPT-4o}$ Compared with Novelty Descriptions from the Paper Introduction and Human Reviewer Evaluations.}
  \label{fig:sam3}
\end{figure*}
\begin{figure*}[ht]
  \includegraphics[width=1.0\linewidth]{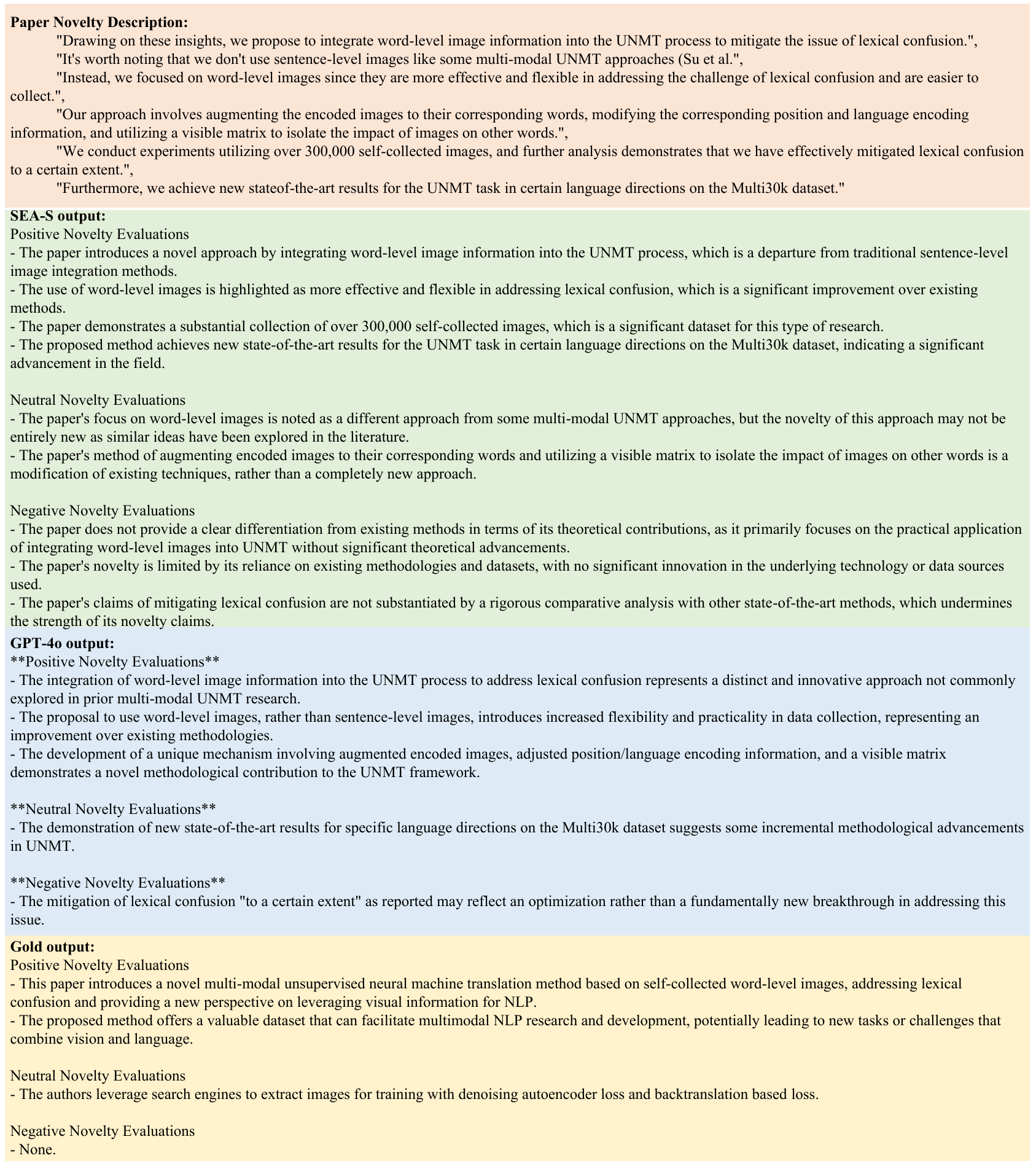}
  \caption{Case Outputs of $\text{SEA-S}$ and $\text{GPT-4o}$ Compared with Novelty Descriptions from the Paper Introduction and Human Reviewer Evaluations.}
  \label{fig:sam4}
\end{figure*}
\begin{figure*}[ht]
  \includegraphics[width=1.0\linewidth]{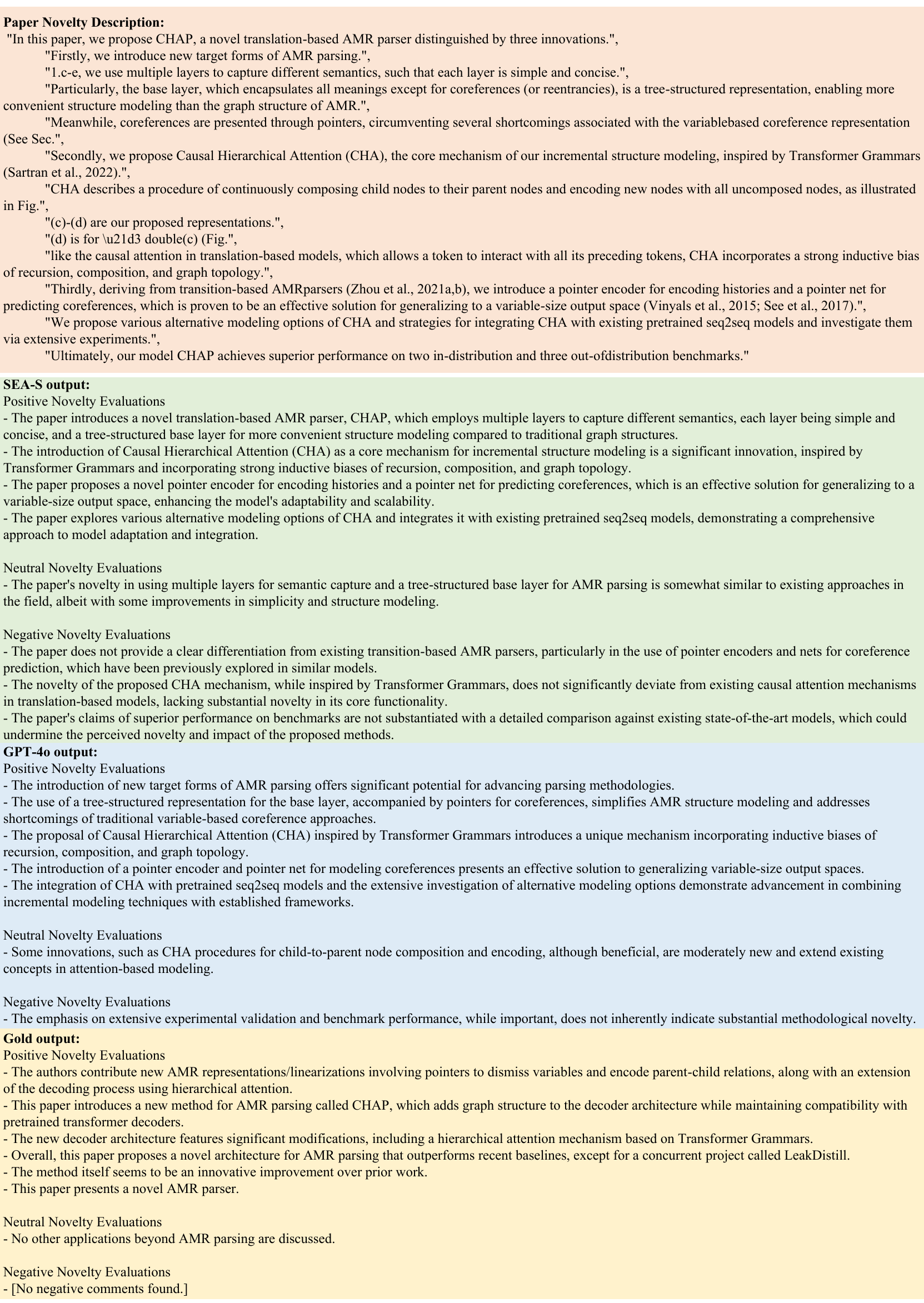}
  \caption{Case Outputs of $\text{SEA-S}$ and $\text{GPT-4o}$ Compared with Novelty Descriptions from the Paper Introduction and Human Reviewer Evaluations.}
  \label{fig:sam5}
\end{figure*}
\section{Case Studies Comparing Human and LLM-Generated Novelty Evaluations}
\label{sec:comp}
We selected five case studies for the analysis presented in Section \ref{sec5.3}. Each case study comprises the novelty description extracted from the paper introduction, the corresponding novelty evaluation provided by the human reviewer, and the novelty evaluations generated by $\text{GPT-4o}$ and $\text{SEA-S}$. These examples are specifically illustrated in Figures \ref{fig:sam1}, \ref{fig:sam2}, \ref{fig:sam3}, \ref{fig:sam4}, and \ref{fig:sam5}.
\section{Additional Analyses}
\label{sec:appendix_analysis}
\subsection{Memorization and Temporal Analysis}
To assess potential data contamination and temporal leakage, we conduct a series of complementary experiments. 
\begin{table}[t]
\centering
\small
\begin{tabular}{lm{0.5cm}<{\centering}m{0.5cm}<{\centering}m{0.5cm}<{\centering}m{0.8cm}<{\centering}}
\toprule
Model & Rel. & Cov. & Clarity & DistAcc \\
\midrule
GPT-3.5 Zero & 3.556 & 0.228 & 0.663 & 0.676 \\
GPT-3.5 Few & 3.505 & 0.246 & 0.660 & 0.731 \\
GPT-3.5 RAG & 3.462 & 0.237 & 0.667 & 0.679 \\
\midrule
GPT-4o Zero & 3.698 & 0.233 & 0.660 & 0.698 \\
GPT-4o Few & 3.561 & 0.240 & 0.659 & 0.709 \\
GPT-4o RAG & 3.448 & 0.224 & 0.667 & 0.697 \\
\midrule
Gemini-2.5-flash Zero & 3.471 & 0.212 & 0.641 & 0.601 \\
Gemini-2.5-flash Few & 3.473 & 0.236 & 0.657 & 0.659 \\
Gemini-2.5-flash RAG & 3.509 & 0.227 & 0.668 & 0.592 \\
\bottomrule
\end{tabular}
\caption{Results of GPT-3.5, GPT-4o and Gemini-2.5-flash on EMNLP 2023.}
\label{tab:gpt35}
\end{table}
First, we evaluate an earlier model (GPT-3.5), released prior to EMNLP 2023, under the same prompting settings as other models. The results (Table~\ref{tab:gpt35}) show that GPT-3.5 performs competitively among general LLMs, indicating that performance is not primarily driven by access to more recent training data.
\begin{table}[t]
\centering
\small
\begin{tabular}{lcccc}
\toprule
Model & Rel. & Cov. & Clarity & DistAcc \\
\midrule
GPT-3.5 Zero & 3.611 & 0.098 & 0.664 & 0.578 \\
GPT-3.5 Few & 3.580 & 0.152 & 0.661 & 0.623 \\
\midrule
GPT-4o Zero & 3.762 & 0.111 & 0.659 & 0.583 \\
GPT-4o Few & 3.580 & 0.118 & 0.658 & 0.555 \\
\bottomrule
\end{tabular}
\caption{Results of GPT-3.5 and GPT-4o on COLING 2020.}
\label{tab:coling}
\end{table}
Second, we perform cross-year evaluation by comparing model performance on COLING 2020 and EMNLP 2023 datasets (Table~\ref{tab:coling}). The results show no substantial performance differences across publication years.

Third, we test for verbatim memorization by prompting models to continue review sentences. In all cases, models respond that they are not certain about the continuation, suggesting the absence of exact recall.
\begin{table}[t]
\centering
\small
\begin{tabular}{lm{0.5cm}<{\centering}m{0.5cm}<{\centering}m{0.5cm}<{\centering}m{0.8cm}}
\toprule
Setting & Rel. & Cov. & Clarity & DistAcc \\
\midrule
GPT-4o Few & 3.569 & 0.221 & 0.659 & 0.699 \\
GPT-4o Few (change) & 3.480 & 0.238 & 0.661 & 0.675 \\
GPT-4o Few (del) & 3.466 & 0.201 & 0.658 & 0.671 \\
\midrule
GPT-4o RAG & 3.460 & 0.243 & 0.665 & 0.684 \\
GPT-4o RAG (change) & 3.430 & 0.196 & 0.667 & 0.681 \\
GPT-4o RAG (del) & 3.392 & 0.168 & 0.668 & 0.687 \\
\midrule
GPT-4o Zero & 3.702 & 0.222 & 0.659 & 0.680 \\
GPT-4o Zero (change) & 3.657 & 0.197 & 0.657 & 0.677 \\
GPT-4o Zero (del) & 3.614 & 0.184 & 0.658 & 0.637 \\
\bottomrule
\end{tabular}
\caption{Results of perturbation experiments on GPT-4o.}
\label{tab:perturb}
\end{table}
Finally, we conduct input perturbation experiments by modifying novelty descriptions through paraphrasing (“change”) and partial deletion (“del”). As shown in Table~\ref{tab:perturb}, model performance remains largely stable across all evaluation dimensions.

Overall, these results consistently suggest that model behavior is not explained by memorization or temporal leakage, but reflects the intrinsic difficulty of novelty evaluation.
\subsection{Analysis by Paper Type}
To investigate whether model performance varies across paper types, we classify papers into coarse-grained categories (methodological and resource papers) using GPT-4o based on titles and abstracts. The benchmark results are then grouped accordingly. Specifically, we report a subset of representative models selected from the main results, including several top-performing models, which sufficiently reflect the overall trends.

\begin{table}[t]
\centering
\small
\begin{tabular}{lcccc}
\toprule
Model & Rel. & Cov. & Clarity & DistAcc \\
\midrule

SEA-E & 3.4234 & 0.2507 & 0.6495 & 0.6869 \\
SEA-S & 3.6270 & 0.2445 & 0.6622 & 0.7194 \\
GPT-4o & 3.6899 & 0.2233 & 0.6599 & 0.6947 \\
Gemini-2.5-flash & 3.4647 & 0.1976 & 0.6409 & 0.5962 \\
\bottomrule
\end{tabular}
\caption{Results on methodological papers.}
\label{tab:method}
\end{table}

\begin{table}[H]
\centering
\small
\begin{tabular}{lcccc}
\toprule
Model & Rel. & Cov. & Clarity & DistAcc \\
\midrule

SEA-E & 3.4337 & 0.2974 & 0.6505 & 0.6712 \\
SEA-S & 3.6402 & \textbf{0.3025} & \textbf{0.6656} & 0.7048 \\
GPT-4o & \textbf{3.7266} & 0.2668 & 0.6582 & \textbf{0.7094} \\
Gemini-2.5-flash & 3.4900 & 0.2608 & 0.6429 & 0.6187 \\
\bottomrule
\end{tabular}
\caption{Results on resource papers.}
\label{tab:resource}
\end{table}
Tables~\ref{tab:method} and~\ref{tab:resource} report the results for methodological and resource papers, respectively. Models consistently achieve better performance on resource papers than on methodological papers. This is likely because resource papers (e.g., benchmarks) present more explicit and concrete contributions, whereas methodological papers often require more nuanced reasoning to assess novelty.

These findings indicate that paper characteristics affect evaluation difficulty, while model rankings remain broadly consistent across categories.
\begin{table}[b]
\centering
\small
\begin{tabular}{lcccc}
\toprule
Model & Mode & High & Low & Diff \\
\midrule
SEA-S & Zero & 0.6761 & 0.6488 & 0.0274 \\
SEA-S & Few & 0.6579 & 0.6351 & 0.0228 \\
SEA-S & RAG & 0.6860 & 0.6474 & 0.0387 \\
GPT-4o & Zero & 0.6632 & 0.6322 & 0.0310 \\
GPT-4o & Few & 0.6615 & 0.6404 & 0.0211 \\
GPT-4o & RAG & 0.6964 & 0.6535 & 0.0428 \\
SEA-E & Zero & 0.6597 & 0.6456 & 0.0141 \\
SEA-E & RAG & 0.6803 & 0.6506 & 0.0297 \\
GPT-5 & RAG & 0.6537 & 0.6108 & 0.0429 \\
Gemini-2.5-flash & RAG & 0.6755 & 0.6338 & 0.0417 \\
\bottomrule
\end{tabular}
\caption{Similarity of LLM-generated evaluations to high- and low-confidence reviews under disagreement.}
\label{tab:confidence}
\end{table}
\subsection{Alignment under Reviewer Disagreement}
We analyze model behavior under reviewer disagreement by examining whether LLM-generated evaluations align differently with reviewers of varying confidence levels. Specifically, we report a subset of representative models selected from the main results, including several top-performing models, which sufficiently reflect the overall trends.

We select samples with substantial disagreement (confidence gap $\geq$ 3) and divide reviews into high-confidence and low-confidence groups. We then compute the semantic similarity between LLM-generated evaluations and each group.

As shown in Table~\ref{tab:confidence}, models consistently exhibit higher similarity to high-confidence reviews. This suggests that LLM-generated evaluations tend to align more closely with reviewers who express stronger certainty, rather than behaving arbitrarily under disagreement.
\end{document}